\RequirePackage{fix-cm}
\RequirePackage{amsmath}
\documentclass[twocolumn,keeplastbox]{svjour3}          % twocolumn
\smartqed  % flush right qed marks, e.g. at end of proof
\usepackage{cite}
\usepackage{graphicx}

\usepackage{tikz}
\usepackage{epstopdf}
\tikzset{every picture/.style={font issue=\footnotesize},
         font issue/.style={execute at begin picture={#1\selectfont}}
        }
\usetikzlibrary{arrows,shapes,positioning}
\usetikzlibrary{patterns}
\usepackage{pgfplots}
\pgfplotsset{compat=1.9}

\usepackage{amssymb}
\usepackage{amsmath,bm}
\usepackage[binary-units=true]{siunitx}
\DeclareMathOperator{\sgn}{sgn}
\DeclareMathOperator*{\argmin}{arg\,min}

\usepackage[linesnumbered,ruled]{algorithm2e}

\usepackage{placeins}

\usepackage[caption=false]{subfig}

\usepackage{multirow}
\usepackage{arydshln}
\usepackage{booktabs,tabularx}
\usepackage{array}

\usepackage[colorlinks=true]{hyperref}

\usepackage{xcolor}
\usepackage{subfiles}
\usepackage{flushend} %remember to add keeplastbox to documentclass

\newcommand{\minitab}[2][l]{\begin{tabular}{#1}#2\end{tabular}}
\begin{document}
\sloppy % solves overflowing margins

\title{Simultaneous multi-descent regression and feature learning for facial landmarking in depth images%\thanks{Grants or other notes
}

\titlerunning{Simultaneous regression and feature learning for facial landmarking}        % if too long for running head

\author {Janez Kri\v{z}aj \and
        Peter Peer \and
        Vitomir \v{S}truc \and
        Simon Dobri\v{s}ek
        %etc.
}


\institute{J. Kri\v{z}aj, V. \v{S}truc, and S. Dobri\v{s}ek\at
           University of Ljubljana, Faculty of Electrical Engineering\\
           Tr\v{z}a\v{s}ka cesta 25, 1000 Ljubljana\\
           Tel.: +386-1-4768-839\\
           Fax: +386-1-4768-316\\
           \email{janez.krizaj@fe.uni-lj.si}           %  \\
           \and
           P. Peer \at
           University of Ljubljana, Faculty of Computer and Information Science\\
           Ve\v{c}na pot 113, 1000 Ljubljana\\
}

\date{Received: date / Accepted: date}

\maketitle

\begin{abstract}
Face alignment (or facial landmarking) is an important task in many face-related applications, ranging from registration, tracking and animation to higher-level classification problems such as face, expression or attribute recognition. While several solutions have been presented in the literature for this task so far, reliably locating salient facial features across a wide range of posses still remains challenging. To address this issue, we propose in this paper a novel method for automatic facial landmark localization in 3D face data designed specifically to address appearance variability caused by significant pose variations. Our method builds on recent cascaded-regression-based methods to facial landmarking and uses a gating mechanism to incorporate multiple linear cascaded regression models each trained for a limited range of poses into a single powerful landmarking model capable of processing arbitrary posed input data. We develop two distinct approaches around the proposed gating mechanism: \textit{i)} the first uses a gated multiple ridge descent (GRID) mechanism in conjunction with established (hand-crafted) HOG features for face alignment and achieves state-of-the-art landmarking performance across a wide range of facial poses, \textit{ii)} the second simultaneously learns multiple-descent directions as well as binary features (SMUF) that are optimal for the alignment tasks and in addition to competitive landmarking results also ensures extremely rapid processing. We evaluate both approaches in rigorous experiments on several popular datasets of 3D face images, i.e., the FRGCv2 and Bosphorus 3D Face datasets and  image collections F and G from the University of Notre Dame. The results of our evaluation show that both approaches are competitive in comparison to the state-of-the-art, while exhibiting considerable robustness to pose variations.               
\end{abstract}

\section{Introduction}
\label{intro}

Face alignment or facial landmarking refers to the task of locating salient facial features in facial images, which is of paramount importance in various applications including face registration and recognition~\cite{Masi18,Wu19}, expression recognition~\cite{Zhao16}, face tracking~\cite{Sanchez18}, normalization of facial pose, size and expressions~\cite{Liu18}, face synthesis from morphable models and facial animation~\cite{Park18} to name a few. In real-world scenarios where face images are often acquired in uncontrolled conditions, one has to deal with various unfavorable factors that adversely affect landmarking performance including pose, expression, and illumination variations as well as partial occlusions of the facial areas. These factors influence the appearance of the facial features in traditional 2D images, e.g.,~\cite{Johnston18} but also in 3D (or better said 2.5D) face data used in this work\footnote{To be strict, we consider 2.5D images in this work, but will use the terms 3D images and depth images interchangeably to refer to this type of data throughout the paper.}. Although some of the existing landmark localization procedures promise to be (at least partially) robust to some of the  factors mentioned above (e.g.~\cite{Perakis13,Sukno15,Feng15}), reliable localization of facial landmarks in the presence of highly variable nuisance factors still remains a considerable challenge.

With the advancement of 3D acquisition technology, landmark localization on 3D facial data has recently been researched extensively ~\cite{Kendrick18,Sukno15}. Many of the 3D landmarking techniques proposed in the literature in the last few years rely on the so-called cascaded-regression framework, where facial landmarks are estimated by regressing from facial features to landmark locations in a cascaded (iterative) manner~\cite{Camgoz15}. Techniques following this framework made considerable advancements towards robust facial landmarking, although they generally still use hand-crafted features, such as SIFTs or HOGs~\cite{Cao15,Camgoz15,Wang17}. Additionally, these methods typically rely on a single regression model in each stage of the cascade to estimate the facial landmarks regardless of the facial characteristics. However, as facial appearance is a complex function of various factors, such as facial shape, pose, incident illumination, expression, and occlusion, a single model is often not sufficient to capture the broad range of variability commonly encountered with facial-image data and to robustly estimate the location of the most salient facial features. 

\begin{figure}[t]
\centering
\null\hfill
\subfloat{\includegraphics[scale=.36]{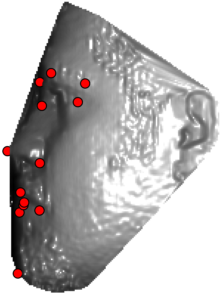}}
\hfill
\subfloat{\includegraphics[scale=.35]{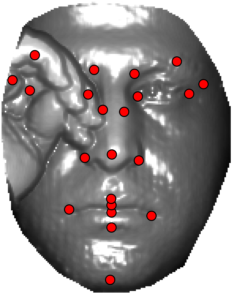}}
\hfill
\subfloat{\includegraphics[scale=.35]{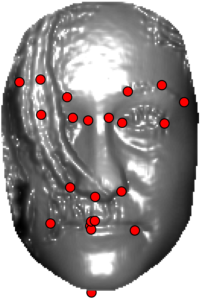}}
\hfill
\subfloat{\includegraphics[scale=.36]{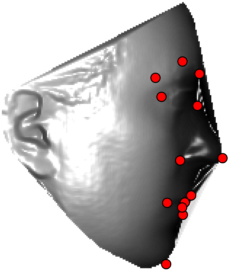}}
\hfill\null
\\[-3ex]
\null\hfill
\subfloat{\includegraphics[scale=.39]{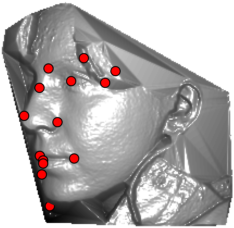}}
\hfill
\subfloat{\includegraphics[scale=.35]{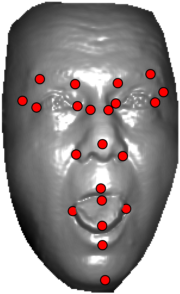}}
\hfill
\subfloat{\includegraphics[scale=.32]{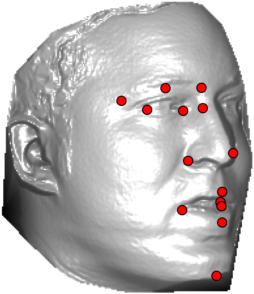}}
\hfill
\subfloat{\includegraphics[scale=.34]{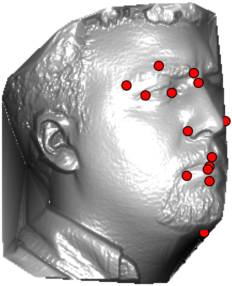}}
\hfill\null
\caption{Sample results of the landmarking GRID approach proposed in this paper. Our model is able to reliably estimate the location of salient facial features in 3D face data even in the presence of large pose variations, i.e., with yaw angles up to $\pm 90^{\circ}$.}\label{fig:rand_results}
\end{figure}

To address this problem, we propose in this paper a novel gating mechanism that incorporates multiple cascaded regression based models each trained for narrow range of facial posses into a single (coherent) landmarking model that is able to reliably estimate the location of salient facial features from arbitrary posed input face data. The combination of simpler view-specific landmarking approaches provides the combined gating-based model the necessary expressive power to describe the considerable appearance variability typically seen with 3D face data captured under different facial poses and consequently allows it to reliably estimate the landmark locations regardless of the facial pose of the input image. The model is partially motivated by the success of earlier methods designed for 2D images that combine multiple landmarking models trained for face alignment of different views, e.g.,~\cite{ramanan2012face, yu2013pose,cootes2000view,li2002multi,zhou2005bayesian}, but unlike these early methods does not rely on parametric appearance or shape models.

We develop two distinct facial landmarking approaches around the proposed gating mechanism. The first relies on a combination of the Gated multiple RIdge Descent (GRID) mechanism and established HOG features and as illustrated in Fig.~\ref{fig:rand_results} achieves remarkable landmarking performance across a broad range of pose variation. Even for poses with yaw rotations of up to $\pm90^{\circ}$, the model is still able to reliably estimate the location of salient facial features. The second approach again relies on the introduced gating mechanism, but in addition to the cascaded regression models needed for face alignment of each group of poses, also learns a feature representation that is used with the regression models for landmark estimation. Specifically, the model Simultaneously learns MUltiple-descent directions as well as binary Feature (SMUF) that are optimal for the alignment task and due to their binary nature also ensure extremely fast execution times. This second approach follows recent trends in computer vision and aims to learn the feature representation that is optimal for a given tasks, but different from deep learning models that are typically used for feature learning \cite{Wang18,Xia19}, uses a computationally much simpler scheme, where binary features are learned based on a learning objective that can be solved using standard optimization procedures.

To evaluate the proposed landmarking approaches, we conduct experiments on multiple datasets of 3D face images, i.e., the Face Recognition Grand Challenge v2 (FRGCv2), the Bosphorus 3D Face datasets and image collections F and G of the University of Notre Dame. We present extensive experiments and comparisons with state-of-the-art methods from the literature. The results of our evaluation show that the GRID ensures state-of-the-art performance for facial landmark localization in 3D face data across pose, while SMUF yields not only competitive landmarking accuracy, but is also extremely fast.   

Our main contributions in this paper are:
\begin{itemize}
\item We propose a gating mechanism for face alignment in 3D face data that allows us to combine multiple alignment models and foster the combined power of the combined models for face alignment across pose. The use of multiple models adds to the overall localization performance, since each model needs to account only for a limited set of plausible facial variations. With this approach reliable landmarking is possible even under large head rotations such as profile facial images, with yaw rotations up to $\pm90^\circ$ (see Fig.~\ref{fig:rand_results}), where many competing methods fail.
\item We develop two distinct landmarking approaches based on the introduced gating mechanism, where one is optimized for performance and the second one is optimized for both performance and speed. We evaluate both approaches in rigorous experiments on multiple face datasets and report competitive performance in comparison to competing methods from the literature. 
\item We study different configurations of the proposed approaches and investigate their behavior when localizing specific facial landmarks.  
\end{itemize}

The rest of the paper is organized as follows: In Section~\ref{sec:rworks} we describe prior work in the field of facial landmarking with the goal of providing the necessary context for our contributions. In Section~\ref{sec:method} we present our gating mechanism and the GRID and SMUF alignment techniques. We describe experiments conducted to evaluate the performance of the proposed methods in discuss results in Section~\ref{sec:exp}. We conclude the paper with some final remarks and future research challenges in Section~\ref{sec:concl}.

\section{Related Work}\label{sec:rworks}

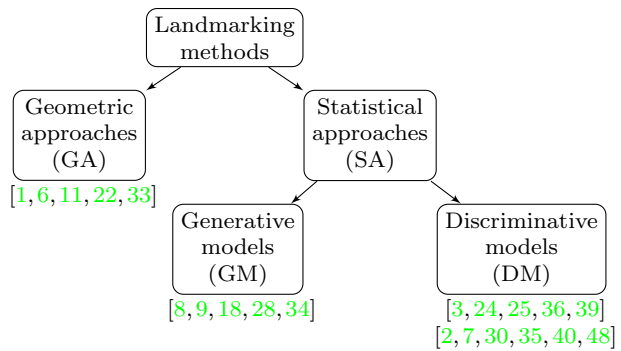
\begin{figure}
\centering
\begin{tikzpicture}
[
block/.style = {rectangle, draw, text centered, rounded corners}, %minimum height=.65cm
line/.style = {draw,-latex'},
node distance = 3mm and 0mm,
auto 
]
\node[block, align=center](methods){Landmarking\\methods};
\node[block,below left= of methods, align=center](geom){Geometric\\approaches\\(GA)};
\node[block,below right= of methods, align=center](stat){Statistical\\approaches\\(SA)};
\node[block, below left= of stat, align=center](gener){Generative\\models\\ (GM)};
\node[block, below right= of stat, align=center](discrim){Discriminative\\models\\ (DM)};
\node[text centered, below= of discrim, align=center,yshift=.35cm](discrim_meths){\cite{Cao15,Sukno15,Wang17,Passalis11,Perakis13}\\\cite{Camgoz15,Zhao11,Romero09,Wang18,Fanelli12,Song14}};
\node[text centered, below= of geom, align=center,yshift=.35cm](geom_meths){\cite{Gupta10,Segundo10,Alyuz10,Mian07,Faltemier08}};
\node[text centered, below= of gener, align=center,yshift=.35cm](gener_meths){\cite{Fanelli13,Smolyanskiy14,Liu14,Feng15,Rai16}};
\path[line](methods)--(geom);
\path[line](methods)--(stat);
\path[line](stat)--(gener);
\path[line](stat)--(discrim);
\end{tikzpicture}
\caption{Taxonomy of 3D landmarking approaches as discussed in the related work section. Recent work is largely focusing on statistical approaches, where landmarking is learned from annotated training data using either generative or discriminative models.}
\label{fig:categ}
\end{figure}

Numerous methods have been proposed in the literature for the task of automatic facial landmark localization over recent years. In this section we present a brief overview of these methods with a focus on alignment techniques that work on 3D images. These techniques can be categorized in various ways, but here we chose to perform a categorization as shown in Fig.~\ref{fig:categ}. We classify existing techniques into two groups: \textit{i}) techniques that are entirely dependent on geometric information and derive prior knowledge about the facial structure and location of facial landmarks by defining a number of heuristic rules and \textit{ii}) techniques that rely on trained statistical models. The latter group of techniques is further divided according to the type of the  model utilized into generative and discriminative methods. A high-level comparison of the related works discussed in this section is given in Table~\ref{tab:methods}.

\begin{table*}
\caption{Summary of the existing 3D facial landmark detection methods}
\label{tab:methods}
\resizebox{\textwidth}{!}{%
\begin{tabular}{lllrr}
\toprule
Author &
Type &
Procedure &
Learning algorithm&
Features
\\\midrule
Mian et al. (2007,~\cite{Mian07})& GA & Nose tip detection through the analysis of horizontal slices & Training-free & Depth data  \\\midrule[0pt]
Faltemier et al.(2008,~\cite{Faltemier08}) & GA & Rotated profile signatures&  Training-free & Rightmost 3D profile lines \\\midrule[0pt]
Gupta et al. (2010,~\cite{Gupta10})&
GA&
ICP coarse alignment, heuristic rules&
Training-free&
Surface curvatures
\\\midrule[0pt]
Segundo et al. (2010,~\cite{Segundo10})&
GA&
Clustering-based face detection, heuristic rules&
Training-free&
Depth relief curves, Surface curvatures
\\\midrule[0pt]
Aly\"{u}z et al. (2010,~\cite{Alyuz10})&
GA&
Facial symmetry axis, heuristic rules&
Training- free&
Shape index, Gaussian curvature
\\\midrule[0pt]
Passalis et al. (2011,~\cite{Passalis11})&
SA-DM&
Candidate landmarks fitting to facial landmark model&
PCA&
Shape index, spin images
\\\midrule[0pt]
Zhao et al. (2011,~\cite{Zhao11})&
SA-DM&
3D statistical facial feature model&
PCA-based learning&
Range map, intensity map
\\\midrule[0pt]
Fanelli et al. (2012,~\cite{Fanelli12})&
SA-DM&
Random forest based voting approach&
Random forests&
Binary tests
\\\midrule[0pt]
Fanelli et al. (2013,~\cite{Fanelli13})&
SA-GM&
Random forests-based regression, AAM&
Random forests&
Binary tests with trees in forest
\\\midrule[0pt]
Perakis et al. (2013,~\cite{Perakis13})&
SA-DM&
Candidate landmark fitting to facial landmark model&
PCA&
Shape index, spin images
\\\midrule[0pt]
Smolyanskiy et al. (2014,~\cite{Smolyanskiy14})&
SA-GM&
2D AMM and 3D morphable face model&
PCA&
RGBD values
\\\midrule[0pt]
Liu et al. (2014,~\cite{Liu14})&
SA-GM&
Normalized cross correlation and depth-based AAM&
PCA&
Shape and depth values
\\\midrule[0pt]
Song et al. (2014,~\cite{Song14})&
SA-GM&
Local coordinate coding&
Coupled dict. learning&
Spin images, synthesized features
\\\midrule[0pt]
Cao et al. (2015,~\cite{Cao15})&
SA-DM&
Cascaded regression&
Linear regression&
Shape indexed and depth features
\\\midrule[0pt]
Sukno et al. (2015,~\cite{Sukno15})&
SA-DM&
Combinatorial search constrained by a flexible shape model&
PCA&
APSC descriptors
\\\midrule[0pt]
Camg\"{o}z et al. (2015,~\cite{Camgoz15})&
 SA-DM&
 Cascaded ridge regression&
 Ridge regression&
 Multi-scale HOG
\\\midrule[0pt]
Feng et al.  (2015,~\cite{Feng15})&
SA-DM&
 Cascaded collaborative  regression&
 Weighted ridge  regression&
 Dynamic multi-scale  HOG
\\\midrule[0pt]
Rai et al.  (2016,~\cite{Rai16})&
SA-GM&
 3D Constrained  Local Models&
 ICA, Point  Dist. model&
 LBP  descriptors
\\\midrule[0pt]
Wang et al.  (2017,~\cite{Wang17})&
SA-DM&
Joint head pose  and facial  landmark regression&
Classif.  guided casc.  regression&
Random forest feature  selection
\\\midrule[0pt]
Liu et al.  (2017~\cite{Liu17})&
SA-DM&
 Hidden Markov  Models (HMMs)&
 HMM  learning&
 Spin  image
\\\midrule[0pt]
Wang et al.  (2018~\cite{Wang18})&
SA-DM&
CNN based feature  extraction and  landmark regression&
Pre-trained CNN  fine-tuning&
CNN-based  global and local  features\\
\bottomrule
\end{tabular}}
\end{table*}

\subsection{Geometric Approaches}

Geometric approaches to facial landmarking are generally training-free and depend solely on the geometric information such as surface curvature or shape index values. A number of rules and heuristics encode the prior knowledge about the relationships between adjacent landmarks (e.g. the nose tip lies on the face symmetry axis, eyes are located above the nose tip, etc.). In most cases, the rules used to define the location of facial landmarks require the face to be in upright and near-frontal position. Moreover, a common downside of these methods is that the landmarks are detected in sequence (commonly starting by detecting a nose tip) and the success rate of finding the next landmark in the sequence is dependent on successfully locating the preceding landmark in the sequence. With these methods an incorrect detection of one landmark affects the detection of all  subsequent landmarks. 

Exemplar geometric methods ~\cite{Gupta10,Segundo10,Alyuz10,Mian07,Faltemier08} start by detecting the nose tip and use its location to constrain the search space of the remaining landmarks. Landmark detection can be grounded on the analysis of Gaussian curvatures~\cite{Gupta10}, profile curvatures~\cite{Faltemier08}, $x$ and $y$ coordinate projections of the depth data~\cite{Segundo10}, shape index values and facial symmetry lines~\cite{Alyuz10} or horizontal slices of range images~\cite{Mian07} to name a few.

\subsection{Statistical Approaches}

Statistical landmarking approaches also exploit local shape information around candidate landmark locations. Additionally, these methods derive some prior knowledge from the training data about the location constraints and encode the acquired knowledge into a statistical model. Thus, these methods require a training set of facial images with annotated landmarks. Unlike training-free geometric approaches, statistics are utilized uniformly for all landmarks, as there is no need for specific rules for each individual landmark. Since all landmarks are handled simultaneously approaches from this group are typically more robust to local distortions, missing data and occlusions of individual landmarks. However, the fact that statistical methods generally address a complete set of landmarks defined by the model could prove to be a problem when a large number of landmarks is (self-)occluded or data is missing from the input images due to acquisition errors. Recently, efforts have been made to handle such problems, e.g.~\cite{Sukno15} use a flexible shape model that works even with an incomplete set of landmarks.

In terms of the type statistical model used, approaches from this group can be divided into techniques that rely either on generative or on discriminative models. We discuss both types of techniques in the following two subsections. 

\subsubsection{Generative Approaches}

Landmark locations can be modeled by generative models, such as Active Appearance Models, Active Shape Models~\cite{Fanelli13,Smolyanskiy14,Liu14} or morphable models~\cite{Feng15}. Techniques from this group often learn face appearances (and/or shape) by conducting Principal Component Analysis (PCA)~\cite{Turk91} on manually annotated and aligned training data. Given a test image, alignment/fitting is achieved by minimizing the difference between the current estimate of the appearance (and/or shape) and the test face. Generally, generative approaches are often computationally expensive and may perform poorly in the presence of occlusions, pose expression and illumination variations due to the involved fitting procedure.

\subsubsection{Discriminative Approaches}

More recent methods to face alignment focus mostly on discriminative approaches that learn a mapping function that predicts the shape, i.e. landmark locations, directly from corresponding image features. Methods from this category  typically offer better landmark localization performance when compared to generative models, especially for faces with greater variability in appearance~\cite{Feng15,Sukno15,Camgoz15,Cao15}. These methods commonly incorporate shape constraints into the models and use local descriptors that are more robust to appearance variations than conventional depth/intensity pixel values used with generative approaches. Discriminative approaches include random forests~\cite{Fanelli12}, graph matching~\cite{Romero09}, cascaded regression~\cite{Camgoz15,Cao15,Feng15,Wang17}, specifically tailored shape models~\cite{Passalis11,Zhao11,Perakis13,Sukno15}, hidden Markov models~\cite{Liu17} and convolutional neural networks~\cite{Wang18}.

The landmarking techniques proposed in this work fall into the group of discriminative approaches. They build on recent face alignment techniques that rely on cascaded regression models that have proven highly successful for landmark localization from 2D face images, e.g.~\cite{Xiong14,Xiong15}. However, compared to these models our solutions exhibit unique features, such as the novel gating mechanism for exploiting multiple pose-specific landmarking models and the ability to incorporate task-specific binary features into the landmarking procedure.    

\section{Methodology}\label{sec:method}

\begin{figure}[!t!]
\includegraphics[trim = 0mm 0mm 0mm 0mm,clip,width=1.0\linewidth]{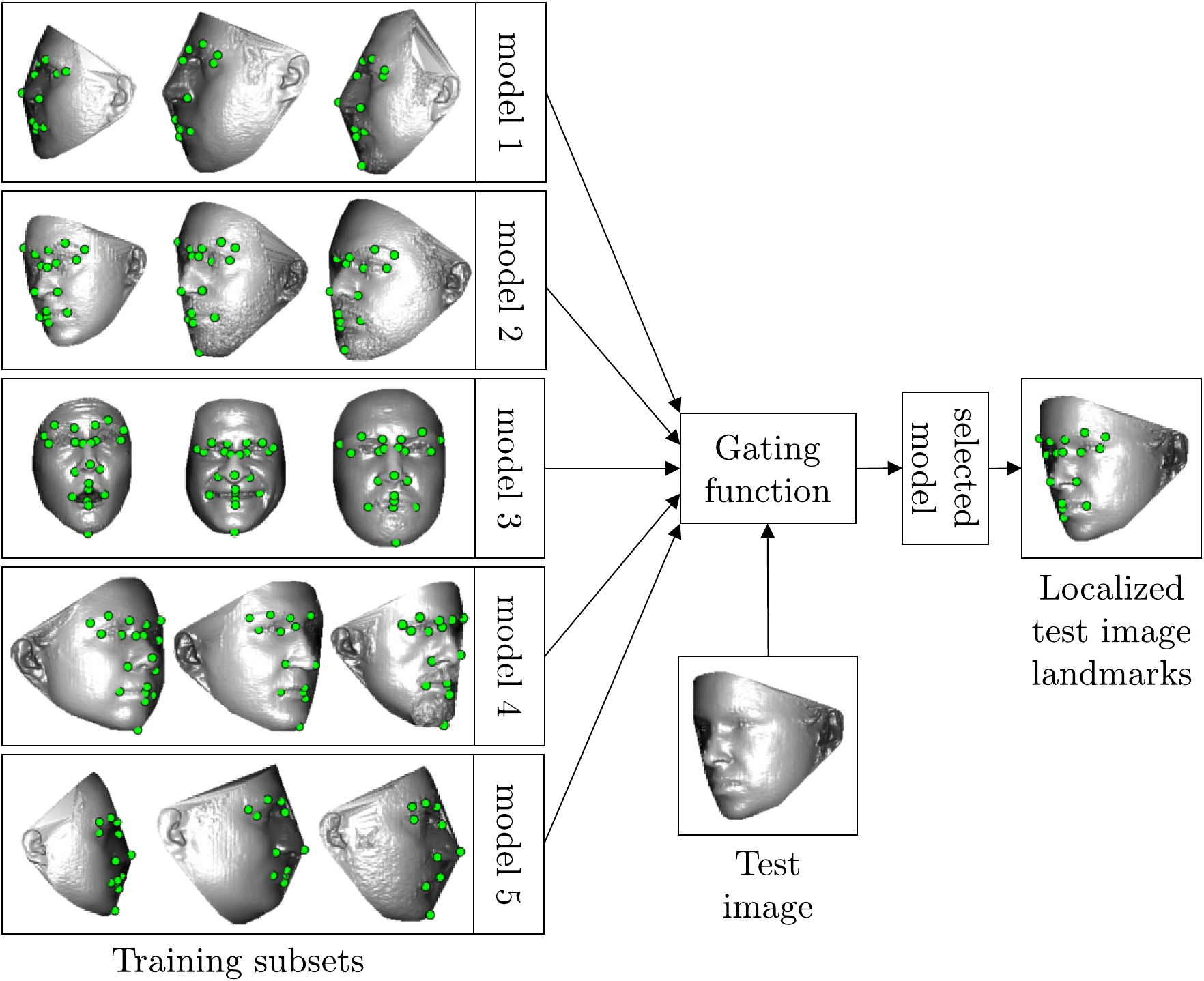}
\caption{Schematic representation of the gating mechanism used in this work. Multiple landmarking models (each containing a cascade of regression models) are trained during the learning stage. At run-time a gating function is used to select the landmarking model that best fits the characteristics of the test data.}
\label{fig:scheme}
\end{figure}

In this section we describe GRID and SMUF, two novel facial landmarking approaches build around the gating mechanism illustrated in Fig.~\ref{fig:scheme}. As can be seen, the gating mechanism partitions the search space for the landmark localization procedure into a number of sub-domains, where each sub-domain encompasses a range of similar facial poses. A separate landmarking model is then trained for each of the sub-domains and the gating mechanism is used to select the most suitable landmarking model for the given tests image. Based on this overall framework, we develop two distinct landmarking techniques, which are described next.

\subsection{GRID Description}

We design GRID (Gated Multiple Ridge Descent) in line with the powerful cascaded regression framework to face alignment, where landmark locations (or in other words, the facial shape) are estimated by regressing from facial features to landmark locations in a cascaded manner. In the first step of this framework, features are extracted from some initial landmark configuration (estimated from the training data) and a regression model is applied on the extracted features to predict landmark updates to better align the landmarks with the actual test image. The update results in a new landmark configuration that forms the basis for the next step in the cascade. The entire procedure is then repeated multiple times and, thus, sequentially refines the predicted locations of the facial landmarks in the test image.  

With GRID, we train multiple cascaded regression models and integrate them into a gated approach that is robust to pose variations. While different regression models and feature representation have been proposed in the literature for facial landmarking, we built GRID around the Supervised Descent Method (SDM) from~\cite{Xiong14} that has not proven successful only for facial landmark localization in 2D images~\cite{Xiong13}, but also for alignment of 3D face images, as we have shown in~\cite{Camgoz15,krivzaj2018localization}.

\subsubsection{Background}

To train the regression models needed for landmarking, SDM requires a number of facial images $\{\bm{I}_n\}_{n=1}^N$, where each image $\bm{I}$ has $L$ landmarks annotated in the form of a shape vector $\bm{x}_*\in\mathbb{R}^{2L\times 1}$. The landmark localization task is then posed as a minimization problem over $\Delta\bm{x}$:
\begin{equation}
\underset{\Delta\bm{x}}{\arg\min}\left\| h(\bm{I},\bm{x}_1+\Delta\bm{x}) - \bm{\phi}_* \right\|^2 \text{,}
\label{eqn:loc_min}
\end{equation}
where $h$ is a feature extraction function, $\bm{\phi}_* = {h}(\bm{I},\bm{x}_*)$ are features extracted around  the  ground  truth  landmarks $\bm{x}_*$,  $\bm{x}_1$ is an initial landmark configuration and $\Delta\bm{x}$ is a landmark update (known for the training data). %Examples of successfully deployed feature extractors $h$ include HOG~\cite{Dalal05} or SIFT~\cite{Lowe04} descriptors. 

Eq.~\eqref{eqn:loc_min} represents a non-linear least squares problem and in general has no closed-form solution. However, it was shown in~\cite{Xiong13} that the problem can be solved through a cascade of least squares regression problems. Thus, for each step $k$ in the cascade, the solution of the least squares problem results in a regression matrix (also referred to as a descent map, DM) $\bm{R}_k$ that can be used to predict the update of the landmark locations from the current image features. The learning algorithm is formulated as a minimization of the loss between the true shape updates $\hat{\bm{x}}_k^n = \bm{x}_k^n-\bm{x}_*^n$ and the expected updates over all training images, i.e.:
\begin{equation}
\underset{\bm{R}_k}{\arg\min}\sum_{n}\left\| \hat{\bm{x}}_k^n - \bm{R}_k\big(\bm{\phi}_{k}^n - \overline{\bm{\phi}_*}\big)\right\|^2 \text{,}
\label{eqn:lls}
\end{equation}
where $n$ is a training-image index and $\overline{\bm{\phi}_*}$ is an average feature vector computed from the ground truth locations $\bm{x}_*^n$ over all training images.  Eq. (\ref{eqn:lls}) now represents a sequence of ordinary least squares regression problems that can be solved in closed form.

During test time, the algorithm starts with some initial landmark locations $\bm{x}_1$, for which the face shape (landmark configuration) is defined by the average landmark locations of training images and the position of the face shape is determined by the face detection procedure\footnote{The average shape is always placed consistently with respect to the detected facial region.}, and sequentially updates the initial estimate to obtain the final landmark locations, i.e.:
\begin{equation}
\bm{x}_{k+1} = \bm{x}_k + \bm{R}_{k}\big(\bm{\phi}_k - \overline{\bm{\phi}_*} \big)\text{,}
\end{equation}
so that the final shape $\bm{x}_k$ converges to $\bm{x}_*$ for all training images. The number of steps $K$ in the cascade, where $k=1,2,...K$ commonly varies depending on the implementation, but usually values of $K$ are between $3$ and $10$

\subsubsection{Ridge Regression}

The original SDM~\cite{Xiong13} formulation uses a least squares solution of Eq.~\eqref{eqn:lls} to learn the DMs, i.e.:
\begin{equation}
\bm{R}_k = \hat{\bm{X}}_k\hat{\bm{\Phi}}_k^\top\big(\hat{\bm{\Phi}}_k \hat{\bm{\Phi}}_k^\top\big)^{-1}\text{,}
\label{eqn:reg}
\end{equation}
where $\hat{\bm{X}}_k$ is a shape matrix with $n$-th column $\hat{\bm{x}}_k^n$ and $\hat{\bm{\Phi}}_k$ is a feature matrix with $n$-th column $\bm{\phi}_{k}^n - \overline{\bm{\phi}_*}$. To solve Eq.~\eqref{eqn:reg} one needs to compute the inverse of $\hat{\bm{\Phi}}_k^\top \hat{\bm{\Phi}}_k$, which, however, may be singular when the size of the feature vectors is too large or when the features are correlated. To overcome this issue, the original SDM applies PCA~\cite{Turk91} to the image features before inverting the matrix.

However, we have shown in~\cite{Camgoz15} that better landmarking performance is achieved if ridge regression is used in the original feature space instead of least squares regression in the PCA subspace. The optimization function in (\ref{eqn:lls}) in this case can be written as
\begin{equation}
\underset{\bm{R}_k}{\arg\min}\sum_{n}\left\| \hat{\bm{x}}_k^n - \bm{R}_k\big(\bm{\phi}_{k}^n - \overline{\bm{\phi}_*}\big)\right\|^2 + \gamma_k\left\|\bm{R}_k\right\|^2\text{,}
\label{eqn:optim_DM}
\end{equation}
where $\gamma_k$ denotes a regularization factor and the solution of Eq.~\eqref{eqn:optim_DM} is computed as:
\begin{equation}
\bm{R}_k = \hat{\bm{X}}_k\hat{\bm{\Phi}}_k^\top\big(\hat{\bm{\Phi}}_k\hat{\bm{\Phi}}_k^\top + \gamma_k\bm{\mathbf{I}}\big)^{-1}\text{,}
\label{eqn:DM}
\end{equation}
where $\bm{\mathbf{I}}$ is an identity matrix. The regularization factor $\gamma_k \geq 0$ controls the general instability of the least squares estimate. Selecting a suitable value for $\gamma_k$, avoids over-fitting and helps to produce estimates of $\bm{R}_k$ that generalize better to unseen data.

\subsubsection{Gated Multiple Ridge Descent}\label{sec:dm_sel}

Experimental results in~\cite{Camgoz15,Xiong15,Xiong13} have shown that the original SDM achieves remarkable landmarking performance on various 2D and 3D datasets. However, it still tends to perform poorly when, for example, large head rotations are present in the facial data~\cite{Xiong15}. Such rotations cause complex facial appearance variations that are difficult to model and hard to account for when using only a single DM in each step of the landmarking cascade. 

To increase the robustness of the model to pose variations, we propose to exploit multiple DMs $\{\bm{R}_k^z\}_{z=\{1:Z\}}$ such that each of the $Z$ DMs accounts for a specific range of head rotations, as illustrated in Fig.~\ref{fig:scheme}. Towards this end, we partition the available training images $\{\bm{I}_n\}_{n=1}^N$ into $Z$ pose-specific subsets and train separate regression cascades for each subset in line with Eq.~\eqref{eqn:DM}.   

\begin{figure}[t]
\centering
\includegraphics[trim = 0mm 0mm 0mm 0mm,clip,width=0.7\linewidth]{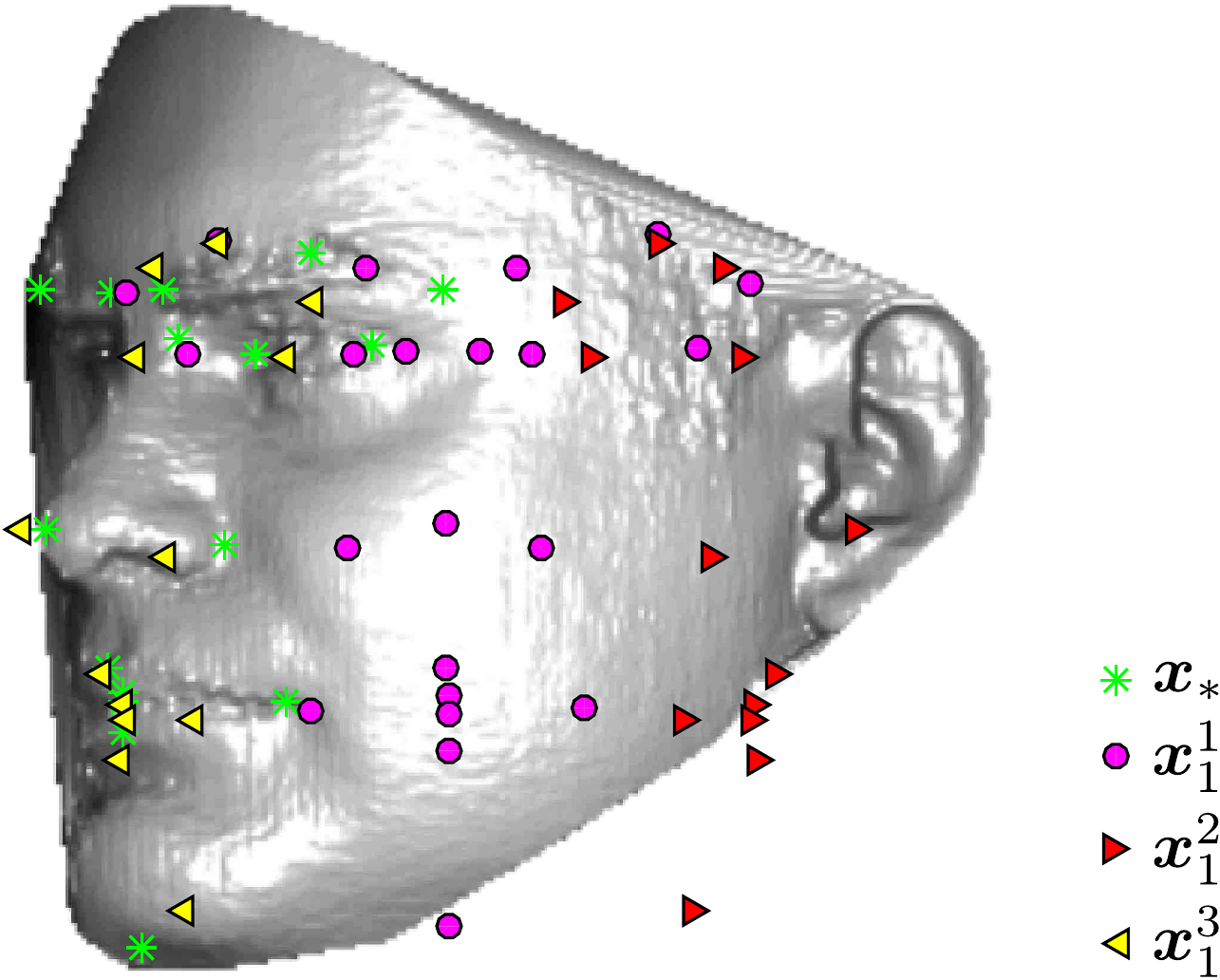}\vspace{1mm}
\caption{Multiple landmark initializations $\{\bm{x}_1^z\}_{z=\{1:3\}}$ and the ground truth landmarks $\bm{x}_*$ superimposed on an example test image. The gating mechanism used in this work determines the pose of  the test image (and consequently selects a landmarking cascade) by comparing features extracted from different shape initializations of the test image to the average features extracted from the true landmark locations of all images in the pose-specific training sets.}
\label{fig:init}
\end{figure}

Once all $Z$ cascades (series of DMs) are trained, a gating function $g_z$ is used to select the most suitable DM (from the $Z$ DMs available in the first cascade stage) for a given test image. The selection procedure begins by computing  features $\{\bm{\phi}_1^z\}_{z=\{1:Z\}}$ from the initial landmark locations $\{\bm{x}_1^z\}_{z=\{1:Z\}}$ in the test image. Here, the initial landmark locations $\bm{x}_1^z$ (see Fig.~\ref{fig:init}) are computed by averaging the ground truth shapes over all training images from the $z$-th training subset:
\begin{equation}
\bm{x}_1^z=\{\overline{\bm{x}_*^n}\}_{n\in z}\text{.}
\end{equation}
The most fitting DM for the given test image %corresponding to the orientation of the test face 
is then selected based on the output of the gating function $g_z$:
\begin{equation}
g_z(\bm{\phi}_1^z) = \sqrt{\frac{1}{m}\big(\bm{\phi}_1^z - \overline{\bm{\phi}_*}\big)^\top \bm{\Sigma}_*^{-1}\big(\bm{\phi}_1^z - \overline{\bm{\phi}_*}\big)}\text{,}
\end{equation}
where $m$ is the feature vector length (in the case of SIFT $m=128\times L$), $\overline{\bm{\phi}_*}$ and $\bm{\Sigma}_*$ are the average and the covariance matrix of the ground truth features over the subset $z$, respectively. We select the subset $z_*$, for which the gating function outputs the lowest value:
\begin{equation}
z_* \in \{1,\ldots, Z\}:g_{z_*}=\min_{z}(g_z)\text{.}
\end{equation}
By doing so, we reliably choose the DM $\bm{R}_1^{z_*}$ that has been trained on images with similar face orientations to the orientation of the test face. For all the subsequent steps $k$ we use the DMs $\bm{R}_k^{z_*}$ that correspond to the $z_*$-th subset  selected in the first step $k=1$ and for efficiency reasons due not change the regression cascade in subsequent steps. Location updates on a given test image are thus computed as
\begin{equation}
\bm{x}_{k+1} = \bm{x}_{k} + \bm{R}_k^{z_*}\big(\bm{\phi}_{k-1}^{z_*} - \overline{\bm{\phi}_*} \big) %\text{.}
\label{eqn:update}
\end{equation}
%\begin{equation}
%\bm{X}_{k+1}^z = \bm{X}_k^z + \bm{R}_k^z\hat{\bm{\Phi}}_k^z\text{.}
%\label{eqn:shape_update}
%\end{equation}

%We use $7$ stages in our cascade ($k=1,2,...,7$), which
The described procedure results in significantly improved performance in the case of large head rotations as shown later in Section~\ref{sec:exp}. 

It needs to be  noted that we rely on HOG features to implement the feature extraction function $h$ in GRID. We select HOG features because of their proven performance in prior landmarking models, e.g.,~\cite{Camgoz15,krivzaj2018localization}.

\subsection{SMUF Description}\label{sec:smuf}

The GRID landmarking approach presented in the previous section relies on HOG features to encode the appearance of the facial landmarks during face alignment. With SMUF (Simultaneous MUlti-descent regression and binary Feature learning) we take a step further and try to learn facial features that are optimal for face alignment. We choose to learn binary features, due to their simplicity and most of all computational simplicity. In the following subsection, we first review the idea of binary feature learning and then develop of SMUF approach that jointly learns a landmarking model as well as corresponding binary features that are optimal for this task. 

\subsubsection{Binary Feature Learning}

Hand-crafted binary features, such as Local Binary Patterns (LBPs)~\cite{pietikainen2011computer} represent powerful image descriptors that have proven highly effective in various computer vision tasks. These features typically rely on pixel comparisons within a local neighborhood and heuristic rules to  encode the pixel comparisons into binary codes. As such, they may be suboptimal and better features could potentially be constructed by learning binary features based on some dedicated learning objective.

\begin{figure}[t]
\centering
\includegraphics[trim = 0mm 0mm 0mm 0mm,clip,width=.9\columnwidth]{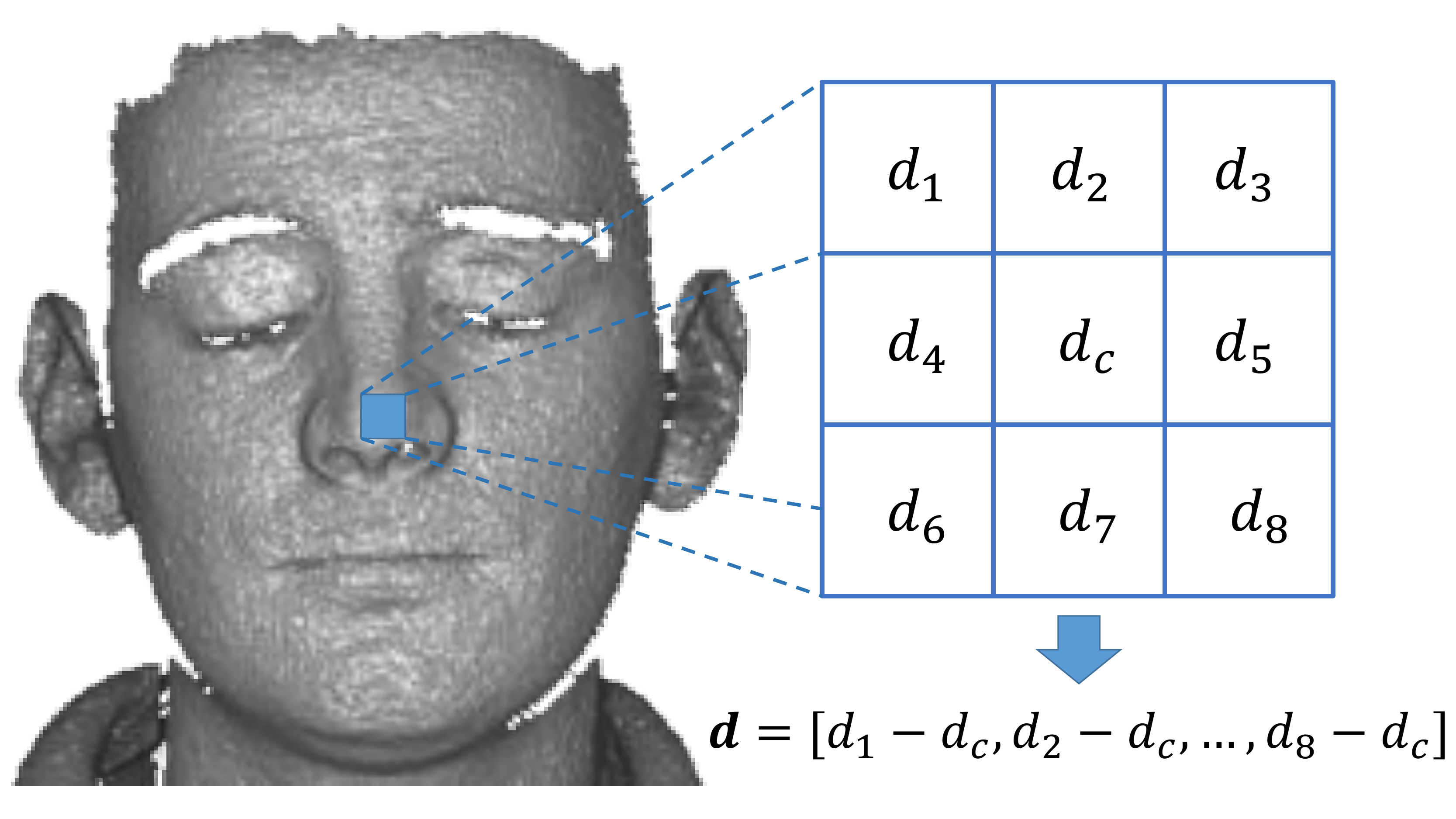}
\caption{We use depth difference vectors $\hat{\bm{d}}$ as the basis for binary feature calculation as proposed in~\cite{lu2015simultaneous,lu2018simultaneous}. The example image above shows how one such vector is computed for a selected landmark. The local pixel neighborhood shown here is of size $3\times 3$ and is selected only for illustration purposes. We use larger neighborhoods for the actual SMUF implementation.}
\label{fig:diff_vector}
\end{figure}

Gong et al.~\cite{gong2013iterative}, for example, propose a learning objective where binary features are learned from an initial image representation $\bm{d}$,   %$\hat{\bm{d}}$
such that the quantization error is minimized. Since binary features $\bm{\phi}$ (containing only $0$s and $1$s) can be computed from $\bm{d}$ %$\hat{\bm{d}}$ 
 as
\begin{equation}
\bm{\phi}=0.5(\sgn(\bm{W}^\top\bm{d})+1)\text{,}
\label{eq: binary learning}
\end{equation}
where $\bm{W}$ is a matrix of hash functions and $\sgn(.)$ stands for the signum function, %that defines the length of the binary code, 
the learning objective $L_q$ that needs to be minimized over $\bm{W}$ on  some training data can be written as:
\begin{equation}
L_q=\left\|{\bm{\phi}} - 0.5 - \bm{W}^\top\hat{\bm{d}} \right\|^2\text{.}
\label{eq:quantization loss}
\end{equation}
It was shown by Lu et al.~\cite{lu2015simultaneous,lu2018simultaneous} that descriptive binary image features can be computed based on the above quantization scheme if pixel (or depth in our case) difference values are used as input $\bm{d}$ %$\hat{\bm{d}}$ 
for binarization. For SMUF we follow this approach and compute one depth difference vector $\bm{d}$ %$\hat{\bm{d}}$ 
for each considered landmark, as illustrated in Fig.~\ref{fig:diff_vector}. 

\subsubsection{Simultaneous DM and Feature Learning}

The learning objective presented in the previous section is focused on representational power, as the binary features are computed in a manner that minimizes a quantization loss. To make the features useful for landmarking, we now formulate a joint optimization function that allows us to simultaneously learn a regression cascade and corresponding binary features that are optimal for the landmarking task. 
%Feature extraction in~(\ref{eqn:DM}) is generally realized by HOG~\cite{Dalal05} or SIFT~\cite{Lowe04}) features, however in this paper we propose a feature learning technique that is incorporated into the landmarking procedure. Generally, local binary features are computationally very efficient and robust to small variations that can be eliminated by quantization~\cite{Lu15}, therefore, we settled to learn binary descriptors directly from depth data.

Let %$\hat{\bm{D}}_k = [\hat{\bm{d}}_k^1,\cdots,\hat{\bm{d}}_k^{LN}]$ 
${\bm{D}}_k = [{\bm{d}}_k^1,\cdots,{\bm{d}}_k^{LN}]$
be a set of depth-difference vectors extracted from patches centered at the facial landmarks $\bm{X}_k = [\bm{x}_k^1,\cdots,\bm{x}_k^{LN}]$ and $k$ stands for the cascade stage, $k=1,2,\cdots,K$. The depth-difference-vector matrix ${\bm{D}}_k$ is mapped to a binary feature matrix $\bm{\Phi}_k$ as follows: 
\begin{equation}
\bm{\Phi}_k=0.5(\sgn(\bm{W}_k^\top{\bm{D}}_k)+1)\text{,}
\end{equation}
where $\bm{W}_k$ is a feature projection matrix and $\sgn(.)$ is again the signum function. To learn $\bm{W}_k$, we formulate the following optimization problem by re-writing~(\ref{eqn:optim_DM}) into matrix form and extending it with the additional constraint $C_2$:
\begin{flalign}
\begin{aligned}
\argmin_{\bm{R}_k,\bm{W}_k} C =&\;C_1 + \lambda C_2 = \\
=&\;\left\| \hat{\bm{X}}_k - \bm{R}_k\tilde{\bm{\Phi}}_k \right\|^2 + \gamma\left\| \bm{R}_k \right\|^2 + \\
&\;+ \lambda\left\| \bm{R_k}(\tilde{\bm{\Phi}}_k - 0.5 - \bm{W}_k^\top\hat{\bm{D}}_k) \right\|^2\text{.}
\end{aligned}&&&
\label{eqn:optimizer}
\end{flalign}
where 
\begin{equation}
\tilde{\bm{\Phi}}_k=0.5(\sgn(\bm{W}_k^\top\hat{\bm{D}}_k)+1)
\label{eqn:binary_feat}
\end{equation}
and $\hat{\bm{D}}_k = \bm{D}_k - \bm{D}_*$, $\bm{D}_*$ are the depth-difference values of the ground truth landmark locations. 
As already emphasized above, the objective of $C_2$ is to minimize the quantization loss between the original depth-difference values and the binarized features, so that most of the depth-difference energy can be preserved in the learned binary features.

We find optimal values for $\bm{R}_k$ and $\bm{W}_k$ by an iterative optimization procedure, where $\bm{W}_k$ is initialized to a random orthogonal matrix. If we assume a fixed $\bm{W}_k$ and compute a partial derivative of $C$ in (\ref{eqn:optimizer}) with respect to $\bm{R}_k$ and set the derivative to zero, we obtain the following solution for $\bm{R}_k$:
\begin{flalign}
\begin{aligned}
&\bm{R}_k = \hat{\bm{X}}_k\tilde{\bm{\Phi}}_k^\top\big[\tilde{\bm{\Phi}}_k\tilde{\bm{\Phi}}_k^\top+\gamma_k\bm{\mathbf{I}} +\\ &\lambda(\tilde{\bm{\Phi}}_k-0.5-\bm{W}_k^\top\hat{\bm{D}}_k)(\tilde{\bm{\Phi}}_k-0.5-\bm{W}_k^\top\hat{\bm{D}}_k)^\top\big]^{-1}\text{.}
\end{aligned}&&&
\label{eqn:update_Rk}
\end{flalign}

In the next step we aim to learn $\bm{W}_k$ with a fixed $\bm{R}_k$ and, hence, rewrite~(\ref{eqn:optimizer}) as follows:
\begin{flalign}
\begin{aligned}
\argmin_{\bm{W}_k} C = &\;\left\| \hat{\bm{X}}_k - \bm{R}_k\bm{W}_k^\top\hat{\bm{D}}_k \right\|^2 + \gamma\left\| \bm{R}_k \right\|^2 +\\
&\;+ \lambda\left\| \bm{R_k}(\tilde{\bm{\Phi}}_k - 0.5 - \bm{W}_k^\top\hat{\bm{D}}_k) \right\|^2\text{.}
\end{aligned}&&&
\label{eqn:optimize_W}
\end{flalign}
If we differentiate~(\ref{eqn:optimize_W}) with respect to $\bm{W_k}$ and set the derivative to zero, we obtain the following update rule for $\bm{W}_k$:
\begin{equation}
\bm{W}_k = \big[(\bm{R}_k^{-1}\hat{\bm{X}}_k+\lambda(\tilde{\bm{\Phi}}_k-0.5))\hat{\bm{D}}_k\big]^\top/(1+\gamma+\lambda)\text{.}
\label{eqn:update_Wk}
\end{equation}
The two optimization steps from \eqref{eqn:update_Rk} and \eqref{eqn:update_Wk} are then repeated until both $\bm{R_k}$ and $\bm{W}_k$ converge. 

Once stable version of $\bm{R_k}$ and $\bm{W}_k$ are obtained, we compute the shape update in accordance with 
\begin{equation}
\bm{X}_{k+1} = \bm{X}_k + \bm{R}_k\tilde{\bm{\Phi}}_k\text{.}
\label{eqn:shape_update}
\end{equation}
and repeat the entire procedure for the next stage in the cascade. Note that because $\tilde{\bm{\Phi}}_k$ is binary, location updates can be computed extremely quickly by simply summing up (specific) rows from $\bm{R}_k$.  

Finally, we compute separate regression cascades and projection matrices for each of the $Z$ training subsets, that is, for each considered group of poses and integrate the computed cascades into the overall SMUF approach using the same gating mechanism as described above for the GRID approach.

%%%%%%%%%%%%%%%%%%%%%%%%%%
%
%Training and testing
%%%%%%%%%%%%%%%%%%%%%%%%%%%
\subsection{Training and testing of GRID and SMUF}

\begin{figure*}
\centering
\includegraphics[trim = 0mm 0mm 0mm 0mm,clip,width=1.0\textwidth]{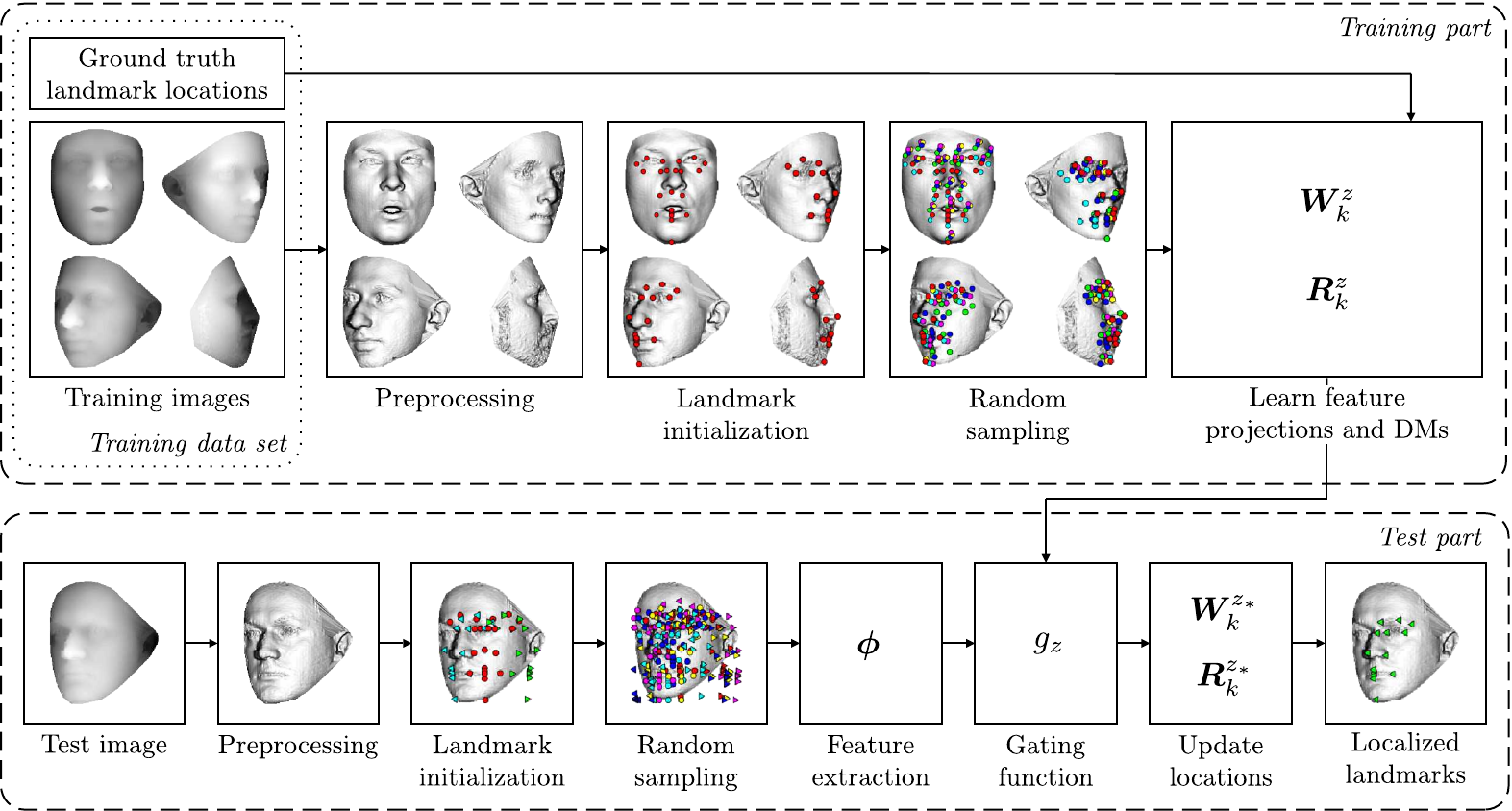}
\caption{Overview of the training and testing stages of the GRID and SMUF landmarking techniques. Both techniques use a similar processing pipeline, but the SMUF approach also learns features (marked by $\bm{W}_k^z)$ in each stage of the training procedure in addition to the landmarking cascade (marked by $\bm{R}_k^z, k=1,2,\ldots,K$) learned by GRID.}
\label{fig:workflow}
\end{figure*}

The overall processing pipeline for the SMUF landmarking approach is shown in Fig.~\ref{fig:workflow}. The procedure for GRID is identical, except for the fact that no features are learned during training. % shows the processing pipeline of the SMDFL landmarking approach. 

The training stage for both methods begins by preprocessing all $N$ training images $\{\bm{I}_n\}_{n=1:N}$ were a depth component of the surface normal is computed in each pixel instead of using original depth values. In each image the face is detected using a simple clustering procedure~\cite{Segundo07} and initial landmark locations $\bm{x}_1^n$ are set based on the detected facial area. To capture the variance of the face detection procedure and to enlarge the amount of training data, we define additional initial landmark locations for each training image by randomly sampling scale and displacement parameters for the detected area from a normal distribution. Starting from the initial locations matrix $\bm{X}_1$ along with the ground truth locations $\bm{X}_*$, a number of DMs $\bm{R}_k^z$ (and for SMUF also projection matrices $\bm{W}_k^z$) are learned. % as summarized in Algorithm~\ref{alg:learning}. 
The updates \eqref{eqn:update_Rk} and \eqref{eqn:update_Wk} are iteratively re-computed till convergence (we empirically estimated that 4 steps are sufficient) for each shape update step $k$ and each of the $Z$ training subsets.

When a test image is presented to the landmarking procedure, it goes through the same face detection, preprocessing, landmark initialization and feature extraction steps as the training images. DMs and feature projections are then selected as described in Section~\ref{sec:dm_sel} and the final landmark locations are computed based on (\ref{eqn:update}).

%\begin{algorithm}
%\SetKwInOut{Input}{Input}
%\SetKwInOut{Output}{Output}
%\caption{SMDFL training procedure}\label{alg:learning}
%\Input{Training images $[\bm{I}_1,\cdots,\bm{I}_N]$ with annotated landmarks $\bm{X}_*=[\bm{x}_*^1,\cdots,\bm{x}_*^N]$ and initial locations $\bm{X}_1=[\bm{x}_1^1,\cdots,\bm{x}_1^N]$.}
%\For{$z=1:Z$}{
%Select $z$-th training subset.\\
%\For{$k=1:K$}{
%%\textbf{Initialization:}\\
%%\Indp
%Initialize $\bm{W}_k^z$ to a random orthogonal matrix.\\
%%\Indm
%%textbf{Optimization:}\\
%\For{$u=1:U$}{
%Extract features $\hat{\bm{\Phi}}_k^z$ according to (\ref{eqn:binary_feat}).\\
%Update $\bm{R}_k^z$ with fixed $\bm{W}_k^z$ using~(\ref{eqn:update_Rkz}).\\
%Update $\bm{W}_k^z$ with fixed $\bm{R}_k^z$ using~(\ref{eqn:update_Wkz}).
%}
%Update shape $\bm{X}_z^k$ according to~(\ref{eqn:shape_update}).\\
%}}
%\Output{Feature projections $\bm{W}_k^z$, descent maps $\bm{R}_k^z$.}
%\end{algorithm}

\section{Experiments}\label{sec:exp}

In this section we evaluate the proposed GRID and SMUF landmarking approaches and compare them to the state-of-the-art. We report landmarking performance in accordance with standard methodology used in this area \cite{Segundo07} for all experiments. Specifically, we use the localization error, i.e. the Euclidean distance in \si{\milli\metre} between the location of the detected landmark and the manually annotated ground truth landmark, for performance reporting. Additionally, we also compute the mean localization error over all landmarks of each test face for some of the experiments. %In some of the tables the results are presented by the mean and standard deviation of the (mean) localization errors over a test data set.

\subsection{Experimental Datasets}
\begin{table}[!t!]
\renewcommand{\arraystretch}{1.2}
\setlength{\tabcolsep}{9.5pt}
\caption{Overview of the datasets used for experimentation. The FRGCv2 dataset is among the most frequently used datasets of 3D face images, whereas the Bosphorus and UND datasets contain challenging images with a high degree of variability in face orientations  and are, hence, well suited for our experiments.}
\label{tab:databases}
\begin{tabular}{lccc}
\toprule
Database&
\#images&
\#subjects&
Variability
\\\midrule
\multirow{2}*{\minitab[@{}l@{}]{FRGCv2}}&
\multirow{2}*{\minitab[@{}c@{}]{4007}}&
\multirow{2}*{\minitab[@{}c@{}]{975}}&
\multirow{2}*{\minitab[@{}c@{}]{Expression}}
\\\\
\multirow{3}*{\minitab[@{}l@{}]{Bosphorus}}&
\multirow{3}*{\minitab[@{}c@{}]{4666}}&
\multirow{3}*{\minitab[@{}c@{}]{105}}&
\multirow{3}*{\minitab[@{}c@{}]{Expression,\\occlusion,\\orientation}}
\\\\\\
\multirow{2}*{\minitab[@{}l@{}]{UND}}&
\multirow{2}*{\minitab[@{}c@{}]{1680}}&
\multirow{2}*{\minitab[@{}c@{}]{537}}&
\multirow{2}*{\minitab[@{}c@{}]{Orientation}}
\\\\\bottomrule
\end{tabular}
\end{table}

We conduct experiments with three popular datasets of 3D face images: the Face Recognition Grand Challenge version 2 (FRGCv2) dataset~\cite{Phillips05}, the Bosphorus 3D Face dataset~\cite{Savran08} and the University of Notre Dame dataset (collections F and G, hereinafter referred to as UND dataset)~\cite{Yan05}. We chose these datasets because they are among the most frequently used 3D face datasets and because they contain challenging 3D images with a high degree of variability in face orientations and are, therefore, well suited for assessing the robustness to such variations. The main characteristics of the datasets are summarized in Table~\ref{tab:databases}.
\begin{table*}[!t!]
\renewcommand{\arraystretch}{1.2}
\caption{Mean localization errors (and standard deviations) for the GRID and SMUF methods on the Bosphorus dataset. Results are reported for different variants of both landmarking techniques implemented with 1, 3, or 5 regression cascades. The best overall performance is achieved with $5$ cascades for both techniques. The results also show that the gating function always selects the correct cascade - observe results for frontal images across the different landmarking variants.}
\label{tab:Bosph}
\centering
{\footnotesize
\begin{tabular}{lrrrrrr}
\toprule
& \multicolumn{6}{c}{Number of descent maps}
\\\cmidrule(l){2-7}
& \multicolumn{2}{c}{1 DM}& \multicolumn{2}{c}{3 DMs}& \multicolumn{2}{c}{5 DMs}
\\\cmidrule(l){2-3}\cmidrule(l){4-5}\cmidrule(l){6-7}
Variation & GRID & SMUF & GRID & SMUF & GRID & SMUF
\\\midrule
Frontal & $3.0\pm1.7$ & $3.1\pm1.8$ & $3.0\pm1.7$ & $3.1\pm1.8$ & $3.0\pm1.7$ & $3.1\pm1.8$\\
Yaw $\leq\pm\ang{10}$ & $3.1\pm1.8$ & $3.2\pm1.9$ & $3.1\pm1.8$ & $3.5\pm3.0$ & $3.1\pm1.8$ & $3.5\pm3.0$\\
Yaw $\leq\pm\ang{20}$ & $3.2\pm1.9$ & $3.6\pm2.5$ & $3.3\pm2.0$ & $3.7\pm2.9$ & $3.3\pm2.0$ &$3.7\pm2.9$\\
Yaw $\leq\pm\ang{30}$ & $3.5\pm2.1$ & $5.2\pm5.1$ & $3.4\pm2.0$ & $3.7\pm2.7$ & $3.4\pm2.0$ & $3.7\pm2.7$\\
Yaw $\leq\pm\ang{45}$ & $5.2\pm4.1$ & $12.1\pm11.4$ & $3.4\pm2.1$ & $3.8\pm2.6$ & $3.4\pm2.1$ & $3.9\pm3.1$\\
Yaw $\leq\pm\ang{90}$ & $9.6\pm10.1$ & $15.6\pm13.1$ & $5.2\pm5.1$ & $7.8\pm8.5$ & $3.5\pm2.1$ & $4.0\pm3.3$\\
Expressions & $3.4\pm2.0$ & $3.6\pm2.3$ & $3.4\pm2.0$ & $3.6\pm2.3$ & $3.4\pm2.0$ & $3.6\pm2.3$\\
Occlusions & $3.9\pm2.5$ & $3.9\pm2.5$ & $3.9\pm2.5$ & $3.9\pm2.5$ & $3.9\pm2.5$ & $3.9\pm2.5$\\
\bottomrule
\end{tabular}}
\end{table*}

The FRGCv2 dataset contains \num{4007} 3D face images of \num{466} individuals. Images of the dataset were acquired with a laser-based Konica Minolta Vivid 910 scanner. Subjects exhibit minor pose variations and various facial expressions. We utilize the ground truth landmarks (\num{8} landmarks per face) from~\cite{Perakis13}, which were manually annotated on a subset of \num{975} images from \num{149} subjects. 

The Bosphorus dataset consists of \num{4666} face samples from \num{105} subjects. Each sample includes a 2D color image, a 3D point cloud and \num{24} manually annotated landmarks (in our experiments we exclude ear dimple landmarks and use the remaining \num{22} landmarks). Next to expression and occlusion variations, images in the dataset also exhibit large variations in pose. Images from the dataset were captured using a structured-light based Inspeck Mega Capturor II Digitizer.

The UND dataset contains \num{1680} semi-profile and profile 3D face images of \num{537} subjects. For our experiments, we use a subset of \num{236} images with yaw rotations of $\pm\ang{45}$ and \num{174} images with yaw rotations of $\pm\ang{60}$ along with the manual annotations (\num{8} landmarks for frontal faces and \num{5} for non-frontal faces) also provided  by~\cite{Perakis13}. Images from this dataset were captured by the same acquisition device as used with FRGCv2.

\subsection{Performance Evaluation on the Bosphorus Dataset}\label{sec:exp_Bosph}

In the first series of experiments we evaluate the performance of GRID and SMUF on the Bosphorus dataset which is particularly suitable to assess the robustness to large pose variations. We perform experiments using a two-fold cross validation setup using half of the images for training and the other half for testing. To increase the size of training data we extend the training set by horizontally flipping each of the aviable training images. We form test sets with respect to the yaw rotation angle or the presence of expressions/occlusions. We implement both methods with $K=7$ cascade stages and use this setup also for all following assessments.

The results of this series of experiments are summarized in Table~\ref{tab:Bosph}. For both GRID and SMUF, we train three landmarking variants, each with a different number of DMs. The first column in Table~\ref{tab:Bosph} marked 1 DM corresponds to the variant that use only one DM that was trained on images with \num{22} annotated landmarks (these are generally images of near frontal faces, since large rotations lead to self-occlusions and fewer annotations). The second column represents the GRID and SMUF variants with \num{3} DMs: one DM is computed in the same way as in the variants in column 1, while the second and the third DM is computed using images with the head rotations up to $\ang{45}$ to the left and right, respectively. The varaints in the third column correspond to the setup in Fig.~\ref{fig:scheme} and contain an additional two DMs corresponding to head rotations in the ranges of $[\ang{45},\ang{90}]$ and $[-\ang{45},-\ang{90}]$. The DM of the near-frontal images are trained using \num{22} landmarks per face image, while the DMs of rotated images are trained using \num{14} landmarks per face as some of the landmarks in these images are typically self-occluded.

As expected, we can observe that the robustness to face rotations is significantly increased when more DMs are utilized. With the GRID and SMUF variants with 5 DMs we achieve reliable landmark localization even on profile face images with yaw rotations up to $\pm\ang{90}$. It can also be seen from the last two rows in Table~\ref{tab:Bosph} that the same localization errors are obtained for all three variants when evaluated on the frontal face images with expression and occlusion variations. This indicates that the expressions and occlusions do not affect the DM selection process since in all cases the frontal DM is correctly chosen by the gating function.

When comparing the performance of SMUF and GRID, we can see that in general GRID ensures slightly better localization results than SMUF for all implemented variants. However, while there is an evident trend towards lower average localization errors for GRID, it is clear from Table~\ref{tab:Bosph} that the performance differences are statistically not significant. Thus, we can conclude that for the Bosphorus dataset both techniques perform more or less equal. %{\color{red}{WHAT CAN WE SEE?}}

\subsection{Evaluation on the FRGCv2 and UND Datasets} 

\begin{figure}
\null\hfill
\subfloat{\includegraphics[width=0.37\linewidth]{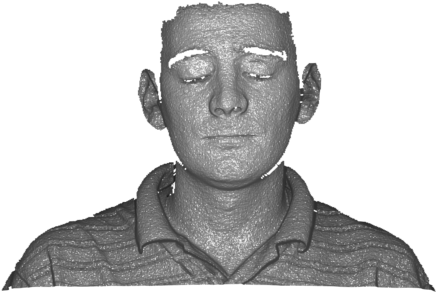}}
\hfill
\subfloat{\includegraphics[width=0.37\linewidth]{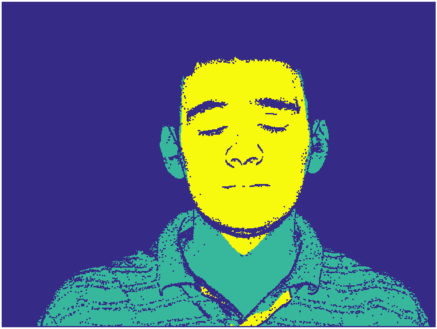}}
\hfill
\subfloat{\includegraphics[height=2.2cm]{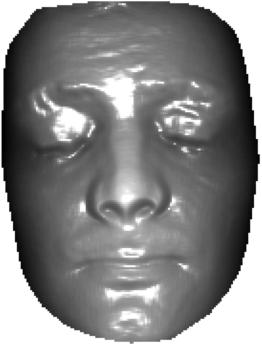}}
\hfill\null
\caption{Illustration of the face detection procedure used on the FRGCv2 and UND datasets. The procedure uses a simple $k$-means clustering approach (with $k=3$) and selects the cluster with the lowest mean depth as the face region. The figure shows: Input image (left), color coded clusters (middle), cropped and smoothed face image (right).}
\label{fig:face_detect}
\end{figure}

In the second series of experiments, we evaluate GRID and SMUF on the joint FRGCv2 and UND datasets. Contrary to the Bosphorus dataset, where images contain solely the head regions and, therefore, using a face detector is not required, images from the FRGCv2 and UND datasets may also contain parts of the upper body and thus face detection is needed to initialize the landmark locations. In this series of experiments we, hence, employ a simple face detector that relies on $k$-means clustering similar to the one presented in~\cite{Segundo07}. Setting the number of clusters to $k=3$ and including several heuristic conditions, this detector divides a 3D image into three regions that most likely correspond to the background, body and head/face regions. The face region is then selected as the cluster with the lowest mean depth value (see Fig.~\ref{fig:face_detect}).

The face detector introduces additional variability into the facial regions, since the detected face may still include smaller parts of the upper body, neck and hair. For that reason we also report face mis-detection rates and selection rates for this experiment. Face mis-detection rate is defined as the percentage of images with the discrepancy between the location of face detection box and the locations of ground truth landmarks. The selection rate is defined as the percentage of images where the correct DM has been selected by the gating function, where we define a DM as incorrect if the DM has been trained on right profile face images while the corresponding test image is facing left, or vice versa. The localization errors are then computed exclusively on the images with correct face detections and DM selections. This type of reporting is adapted from~\cite{Perakis13}, which we use for baseline comparison in this experiment. 

The results of the experiments are presented in Table~\ref{tab:FRGCv2_UND}. For details on the dataset abbreviations in the first column please refer to~\cite{Perakis13}, since the experimental setup and the landmark annotations are adopted from there. In short, however, DB00F denotes an image subset with varying facial expressions, which is further partitioned into neutral (neut.), mild and extreme (extr.) facial expressions. The remaining image subsets contain faces with $45^{\circ}$ or $60^{\circ}$ yaw rotations either to the right (R), the left (L) or both (RL). As the GRID and SMUF methods require non-frontal images to train some of the DMs, our experimental setup differs from~\cite{Perakis13} only in the construction of training set where we also employ images from the Bosphorus dataset.

Detection and selection rates are consistently above $95\%$ for all subsets as it can be observed from the first two columns in Table~\ref{tab:FRGCv2_UND}. Localization errors of our two landmarking approaches  are compared to Perakis et al.~\cite{Perakis13} (last column) which, to the best of our knowledge, achieves the highest performance in the literature on these datasets. The results show the robustness of our methods to both expression variations and to rotations. The mean localization error is under $6$~\si{\milli\metre} on all tested subsets for both GRID and SMUF. Since the training data is taken from the Bosphorus dataset (acquired with a different 3D camera), the results also imply good generalization to data from different sensors. All experiments from this section were performed using 5 DMs, as we observed earlier in Section~\ref{sec:exp_Bosph} that this setting is the most robust to rotation variations.

\begin{comment}
\begin{table}
\renewcommand{\arraystretch}{1.2}
\setlength{\tabcolsep}{4.5pt}
\caption{Mean localization errors (and corresponding standard deviations) on the FRGCv2 and UND datasets. Results are reported for specific subsets in accordance with the protocol from~\cite{Perakis13}. The subsets feature images with different levels of expression variations (DB00F), and yaw rotations of $45^{\circ}$ and $60^{\circ}$ to the right (R), the left (L) or in both directions (RL). The results show that both GRID and SMUF offer competitive performance when compared to the method from~\cite{Perakis13}.   }
\label{tab:FRGCv2_UND}
\begin{tabular}{l@{}rrrr}
\toprule
&\multirow{2}*{\minitab[@{}r@{}]{Detect.\\rate [\%]}}&\multirow{2}*{\minitab[@{}r@{}]{Select.\\rate [\%]}}&\multicolumn{2}{c}{Localization error}\\
\cmidrule(l){4-5}
Data set&&&\textbf{SMDFL}&Perakis~\cite{Perakis13}\\
\midrule
DB00F&$99.59$&$100.00$&$3.20\pm1.70$&$5.00\pm1.85$\\
DB00F-neut.&$99.55$&$100.00$&$2.97\pm1.64$&$4.52\pm1.51$\\
DB00F-mild&$99.72$&$100.00$&$3.39\pm1.71$&$4.95\pm1.46$\\
DB00F-extr.&$99.44$&$100.00$&$3.92\pm1.79$&$6.28\pm2.60$\\
DB00F45RL&$98.29$&$99.13$&$4.81\pm3.07$&$4.97\pm1.92$\\
DB45R&$99.15$&$97.44$&$5.53\pm1.94$&$5.03\pm1.92$\\
DB45L&$98.31$&$99.14$&$4.95\pm1.91$&$4.75\pm1.91$\\
DB60R&$98.85$&$96.55$&$5.35\pm3.85$&$4.95\pm1.80$\\
DB60L&$98.85$&$95.40$&$5.95\pm4.69$&$5.30\pm2.49$\\
\bottomrule
\end{tabular}
\end{table}
\end{comment}

\begin{table}
\renewcommand{\arraystretch}{1.2}
\setlength{\tabcolsep}{3pt}
\caption{Mean localization errors (and corresponding standard deviations) on the FRGCv2 and UND datasets. Results are reported for specific subsets in accordance with the protocol from~\cite{Perakis13}. The subsets feature images with different levels of expression variations (DB00F), and yaw rotations of $45^{\circ}$ and $60^{\circ}$ to the right (R), the left (L) or in both directions (RL). The results show that both GRID and SMUF offer competitive performance when compared to the method of Perakis et al.~\cite{Perakis13}.}
\label{tab:FRGCv2_UND}
\begin{tabular}{lrrrrr}
\toprule
&\multirow{2}*{\minitab[@{}r@{}]{Det.\\rate}}&\multirow{2}*{\minitab[@{}r@{}]{Sel.\\rate }}&\multicolumn{3}{c}{Localization error}
\\\cmidrule(l){4-6}
Data set & & & GRID & SMUF & Perakis\cite{Perakis13}
\\\midrule
DB00F & $99.6$ & $100.0$ & $3.2\pm1.7$ & $3.4\pm2.2$ & $5.0\pm1.9$\\
DB00F-neut. & $99.6$ & $100.0$ & $3.0\pm1.6$ & $3.2\pm1.8$ & $4.5\pm1.5$\\
DB00F-mild & $99.7$ & $100.0$ & $3.3\pm1.6$ & $3.5\pm1.9$ & $5.0\pm1.5$\\
DB00F-extr. & $99.4$ & $100.0$ & $4.0\pm2.1$ & $4.2\pm3.0$ & $6.3\pm2.6$\\
DB00F45RL & $98.3$ & $99.1$ & $3.3\pm2.0$ & $4.0\pm2.9$ & $5.0\pm1.9$\\
DB45R & $99.1$ & $97.4$ & $3.2\pm2.1$ & $3.7\pm2.6$ & $5.0\pm1.9$\\
DB45L & $98.3$ & $99.1$ & $3.8\pm2.1$ & $4.1\pm2.7$ & $4.8\pm1.9$\\
DB60R & $98.9$ & $96.6$ & $3.7\pm1.6$ & $4.2\pm2.5$ & $5.0\pm1.8$\\
DB60L & $98.9$ & $95.4$ & $4.2\pm1.9$ & $5.1\pm4.0$ & $5.3\pm2.5$\\
\bottomrule
\end{tabular}
\end{table}

\begin{table*}[!t!]
\renewcommand{\arraystretch}{1.2}
\setlength{\tabcolsep}{4.5pt}
\caption{Localization errors of GRID and SMUF in comparison to the state-of-the-art on non-frontal facial data sets for 10 common facial landmarks. GRID and SMUF significantly outperform competing methods on all experimental datasets. When comparing the learned binary features from SMUF to handcrafted LBP features we observe better performance for the learned binary features. GRID and SMUF again perform similarly for all landmarks.}
\label{tab:compare}
\centering
\resizebox{\textwidth}{!}{%
\begin{tabular}{lrrrrrrrrr}
\toprule
&
\multicolumn{9}{c}{Method \& database}
\\\cmidrule(l){2-10}
&
\multirow{2}*{\minitab[@{}r@{}]{Sukno\\et al.\cite{Sukno15}}}& \multirow{2}*{\minitab[@{}r@{}]{Creusot\\et al.\cite{Creusot13}}}& 
\multirow{2}*{\minitab[@{}r@{}]{Passalis\\et al.\cite{Passalis11}}}& \multirow{2}*{\minitab[@{}r@{}]{Perakis\\et al.\cite{Perakis13}}}&
\multirow{2}*{\minitab[@{}r@{}]{LBP}}&
\multicolumn{2}{c}{\multirow{2}*{\minitab[@{}r@{}]{GRID}}}&
\multicolumn{2}{c}{\multirow{2}*{\minitab[@{}r@{}]{SMUF}}}
\\\\\cmidrule(l){7-8}\cmidrule(l){9-10}
\multirow{2}*{\minitab[@{}r@{}]{\\Landmarks}}&
\multirow{2}*{\minitab[@{}r@{}]{Bosphorus}}&
\multirow{2}*{\minitab[@{}r@{}]{Bosphorus}}&
\multirow{2}*{\minitab[@{}r@{}]{FRGC\\+UND}}&
\multirow{2}*{\minitab[@{}r@{}]{FRGC\\+UND}}&
\multirow{2}*{\minitab[@{}r@{}]{Bosphorus}}&
\multirow{2}*{\minitab[@{}r@{}]{FRGC\\+UND}}&
\multirow{2}*{\minitab[@{}r@{}]{Bosphorus}}&
\multirow{2}*{\minitab[@{}r@{}]{FRGC\\+UND}}&
\multirow{2}*{\minitab[@{}r@{}]{Bosphorus}}
\\\\\midrule
Inner eye c.&
$2.9\pm2.0$&
$4.1\pm2.6$&
$6.4\pm3.0$&
$4.8\pm2.7$&
$2.8\pm3.4$&
$2.7\pm1.5$&
$2.0\pm1.1$&
$3.2\pm2.1$&
$2.4\pm1.9$
\\
Outer eye c.&
$5.1\pm3.7$&
$6.3\pm4.0$&
$6.6\pm3.7$&
$5.7\pm3.9$&
$3.5\pm3.8$&
$2.8\pm1.9$&
$2.5\pm1.4$&
$3.4\pm2.3$&
$2.9\pm2.8$
\\
Nose tip&
$2.3\pm1.8$&
$4.3\pm2.6$&
$4.6\pm3.0$&
$4.4\pm2.7$&
$4.9\pm6.3$&
$3.5\pm2.8$&
$3.7\pm2.8$&
$4.8\pm4.4$&
$3.9\pm3.5$
\\
Nose c.&
$3.0\pm1.9$&
$4.2\pm2.4$&
n/a&
n/a&
$4.0\pm4.7$&
n/a&
$3.0\pm2.0$&
n/a&
$3.5\pm2.9$
\\
Mouth c.&
$6.1\pm5.1$&
$8.0\pm5.4$&
$5.8\pm3.9$&
$5.0\pm2.9$&
$3.4\pm4.3$&
$2.9\pm2.0$&
$2.5\pm1.6$&
$3.4\pm2.5$&
$2.7\pm2.1$
\\
Chin tip&
$7.6\pm6.7$&
$15.4\pm10.5$&
$6.6\pm3.5$&
$4.8\pm3.5$&
$5.5\pm4.8$&
$4.8\pm2.1$&
$4.8\pm3.4$&
$6.2\pm3.7$&
$5.1\pm3.5$
\\\bottomrule	
\end{tabular}}
\end{table*}

\subsection{Comparison to the State-of-the-art}

In the next series of experiments, we compare the performance of GRID and SMUF to the performance of state-of-the-art landmarking methods from the literature. Specifically, we select the method of Sukno et al.\cite{Sukno15}, the technique of Creusot et al. \cite{Creusot13}, and the landmarking approaches of Passalis et al.\cite{Passalis11} and Perakis et al. \cite{Perakis13} for our comparison. To the best of our knowledge these landmarking methods are the only ones that were evaluated on both frontal as well as rotated 3D facial images. Following the experimental protocols of other authors, we used the DB00F45RL subset when performing experiments on the FRGCv2+UND database and used the entire database for experimentation on the Bosphorus dataset. Additionally, we also implement our gated landmarking approach with hand-crafted binary features, that is, with LBPs (uniform, neighborhood size of 8 and radius of 1) to capitalize on the usefulness of learning binary features instead of using off-the-shelf binary feature extractors.

\begin{figure}[!t!]
\centering
\hfill\null
\subfloat[{\footnotesize GRID, FRGC/UND}]{\includegraphics[height=4.5cm]{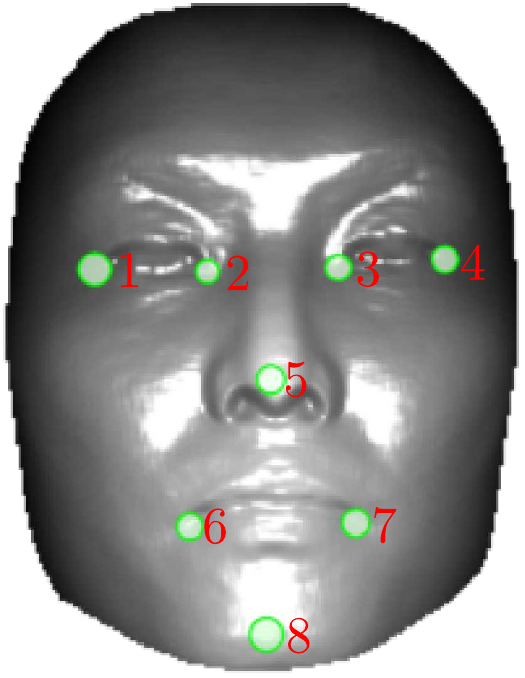}}
\hfill
\subfloat[{\footnotesize SMUF, Bosphorus}]{\includegraphics[height=4.5cm]{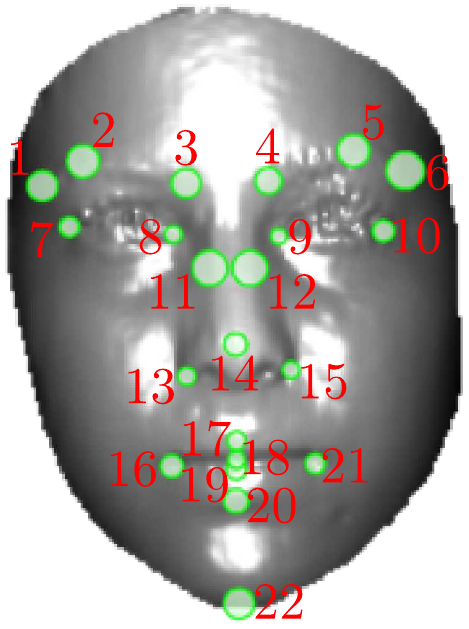}}
\hfill\null 
\caption{Mean localization errors achieved by GRID and SMUF for individual landmarks of the FRGCv2+UND and Bosphorus datasets. The size of the circles corresponds to the localization errors. The numbering of the landmarks as shown here is also used in Figs.~\ref{fig:boxplot_bosph} and \ref{fig:boxplot_und}). The lowest errors are achieved on distinct landmarks with corner-like properties, e.g., the mouth corners.}
\label{fig:lm_var}
\end{figure}

The results of the comparison are shown in Table~\ref{tab:compare}. We observe that on the Bosphorus dataset both GRID and SMUF significantly outperform the competing methods from the literature and achieve not only lower average localization errors, but also significantly smaller standard deviations on these errors. The only exception here are the nose tip and corners, where the method of Sukno performs similarly or slightly better. We also see similar results for the FRGC-UND dataset, where both GRID and SMUF achieve a considerable reduction in the localization errors for all considered landmarks compared to the state-of-the-art. 

When comparing the learned binary features used in SMUF to the hand-crafted LBP features, we also see an obvious performance improvement with the learned features, re-enforcing our assumption that learning binary features is beneficial for face alignment. The comparison between GRID and SMUF shows a similar picture as in the previous series of experiments, where GRID was found to perform slightly better than SMUF, but not significantly so.

\begin{figure*}[!t!]
\centering
\includegraphics[trim = 4mm 2mm 12mm 5mm,clip,width=0.97\linewidth]{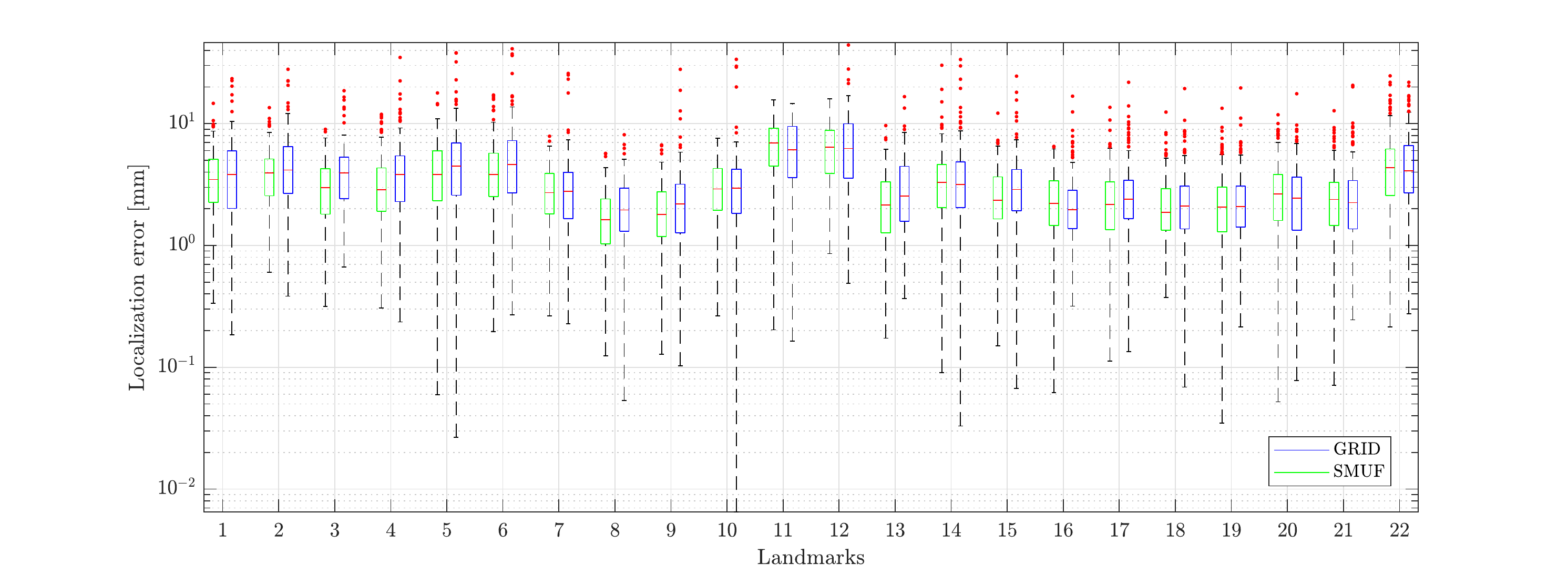}
\caption{Localization errors in the form of box and whiskers plots for the Bosphorus dataset achieved with the GRID and SMUF landmarking techniques. The results show that the lowest localization errors with both methods are achieved on distinct landmarks with corner-like characteristics, such as the eye or mouth corners or the nose tip. The figure is best viewed in color.}
\label{fig:boxplot_bosph}
\end{figure*}

\begin{figure*}[!t!]
\centering
\hfill
\includegraphics[trim = 4mm 2mm 12mm 0mm,clip,width=0.97\columnwidth]{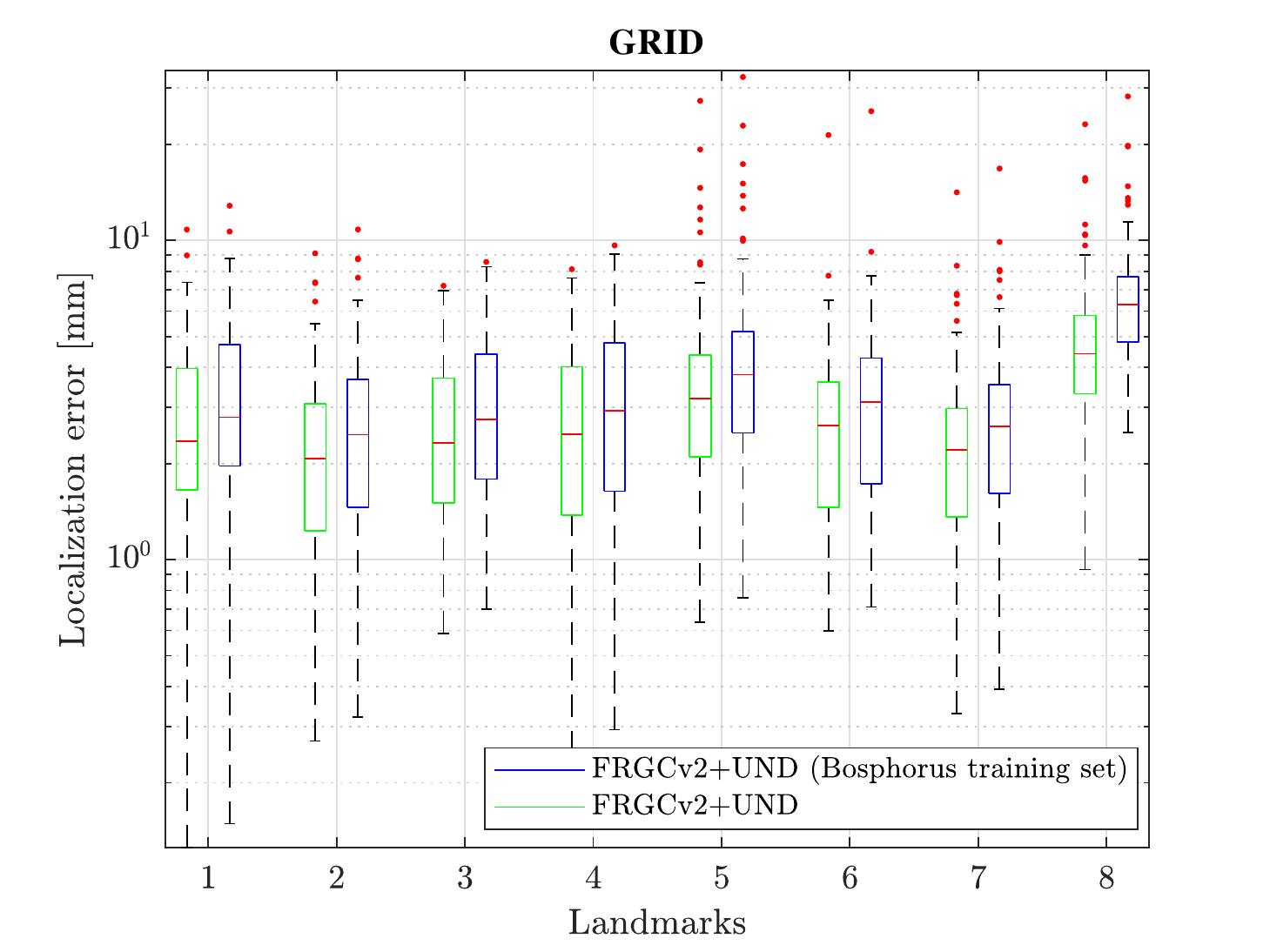}
\hfill
\includegraphics[trim = 4mm 2mm 12mm 0mm,clip,width=0.97\columnwidth]{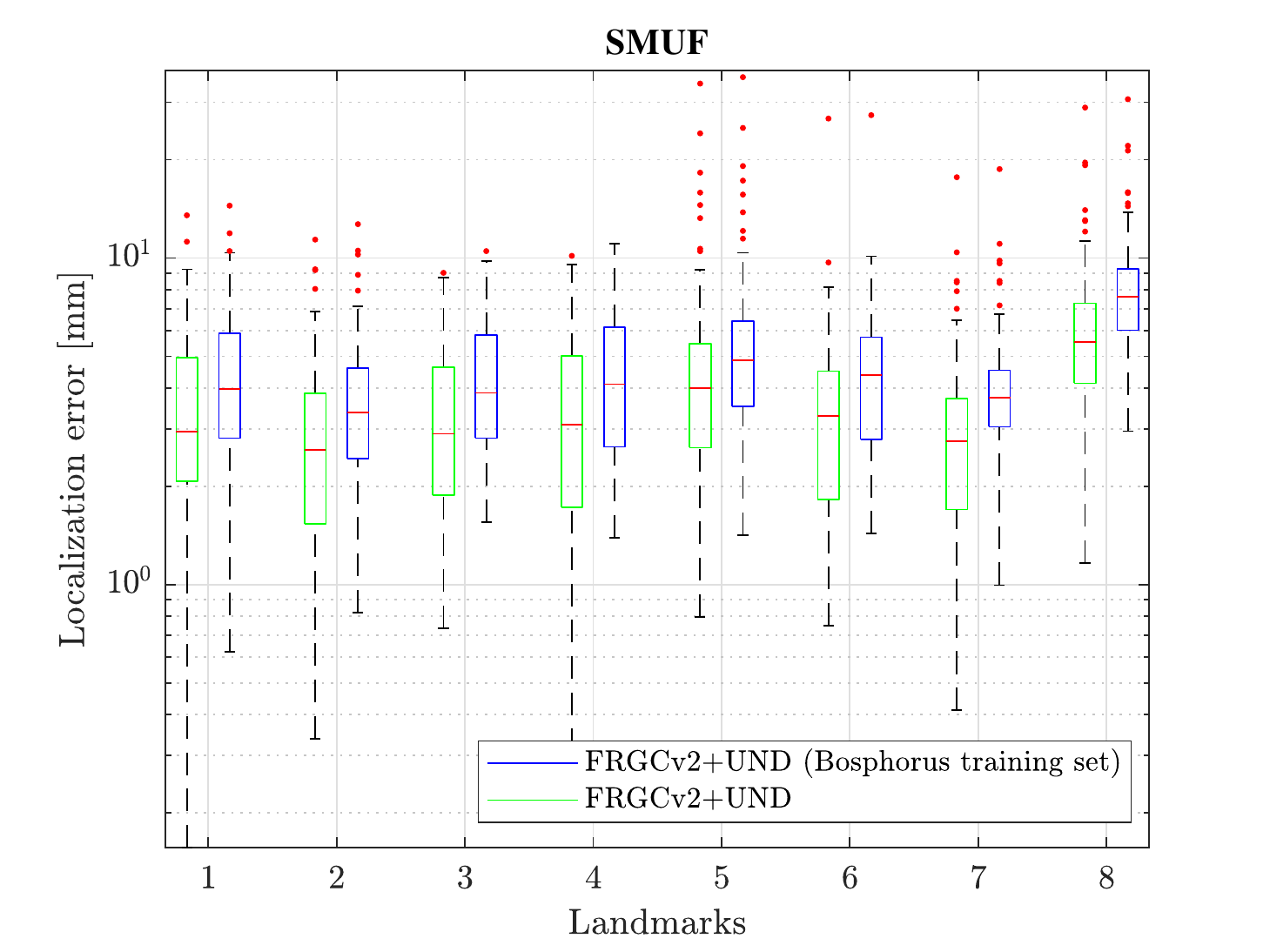}
\hfill
\caption{Localization errors in the form of box and whiskers plots achieved on the FRGCv2+UND dataset with the GRID and SMUF landmarking techniques. Results are also presented for a cross-dataset experiment, where the landmarkers are trained on the Bosphorus datasets and are evaluated on the FRGV2+UND dataset. Lower errors are again achieved on distinct landmarks. The methods generalize well to novel datasets with the median errors for the cross-dataset experiment being only slightly larger that for the within-dataset experiments for the majority of landmarks. The figure is best viewed in color.}
\label{fig:boxplot_und}
\end{figure*}

\begin{figure*}
\centering
\hfill\null
\subfloat[GRID]{\includegraphics[scale=.35]{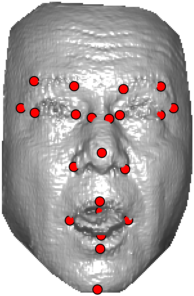}\label{fig:bosph_expr_t1}}
\hfill
\subfloat[SMUF]{\includegraphics[scale=.35]{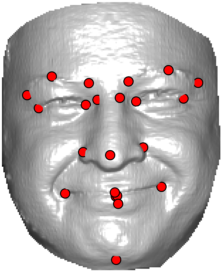}\label{fig:bosph_expr_t2}}
\hfill
\subfloat[GRID]{\includegraphics[scale=.35]{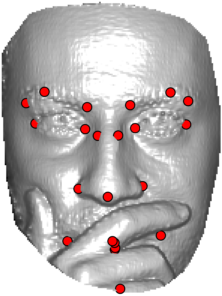}\label{fig:bosph_occl_t1}}
\hfill
\subfloat[SMUF]{\includegraphics[scale=.35]{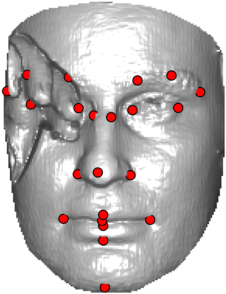}\label{fig:bosph_occl_t2}}
\hfill
\subfloat[GRID]{\includegraphics[scale=.35]{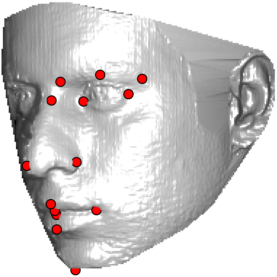}\label{fig:bosph_rot_t1}}
\hfill
\subfloat[SMUF]{\includegraphics[scale=.35]{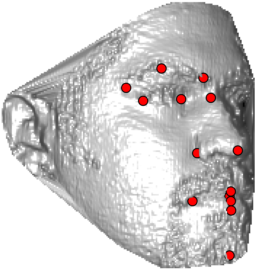}\label{fig:bosph_rot_t2}}
\hfill
\subfloat[GRID]{\includegraphics[scale=.35]{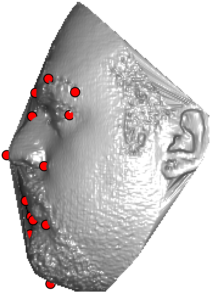}\label{fig:bosph_rot_t3}}
\hfill
\subfloat[SMUF]{\includegraphics[scale=.35]{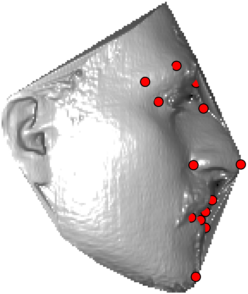}\label{fig:bosph_rot_t4}}
\hfill\null
\\%[-1ex]
\null\hfill
\subfloat[GRID]{\includegraphics[scale=.35]{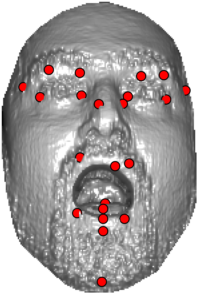}\label{fig:bosph_expr_f1}}
\hfill
\subfloat[SMUF]{\includegraphics[scale=.35]{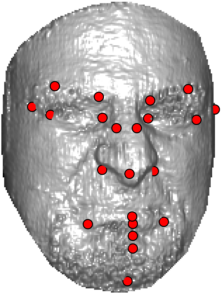}\label{fig:bosph_expr_f2}}
\hfill
\subfloat[GRID]{\includegraphics[scale=.35]{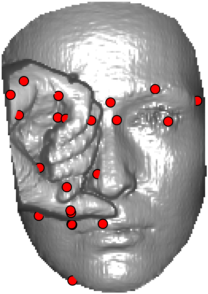}\label{fig:bosph_occl_f1}}
\hfill
\subfloat[SMUF]{\includegraphics[scale=.35]{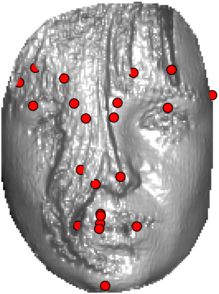}\label{fig:bosph_occl_f2}}
\hfill
\subfloat[GRID]{\includegraphics[scale=.35]{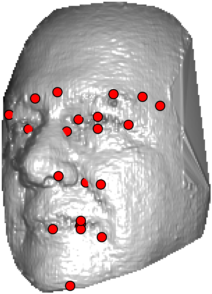}\label{fig:bosph_rot_f1}}
\hfill
\subfloat[SMUF]{\includegraphics[scale=.35]{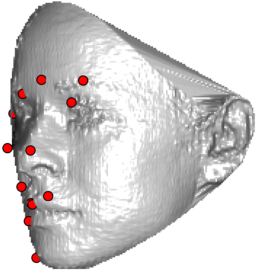}\label{fig:bosph_rot_f2}}
\hfill
\subfloat[GRID]{\includegraphics[scale=.35]{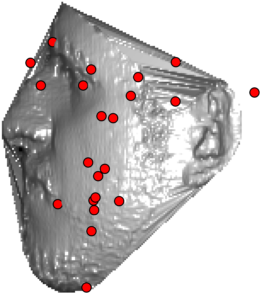}\label{fig:bosph_dm_f1}}
\hfill
\subfloat[SMUF]{\includegraphics[scale=.35]{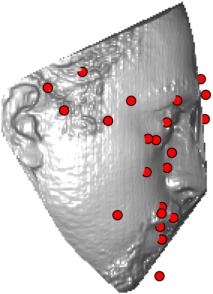}\label{fig:bosph_dm_f2}}
\hfill\null
\caption{Exemplar landmark detection results on the Bosphorus database: the first row depicts randomly chosen test samples~\protect\subref{fig:bosph_expr_t1}-\protect\subref{fig:bosph_rot_t4}, while the second row includes samples with \textbf{high} localization errors due to expressions~\protect\subref{fig:bosph_expr_f1}-\protect\subref{fig:bosph_expr_f2}, occlusions~\protect\subref{fig:bosph_occl_f1}-\protect\subref{fig:bosph_occl_f2}, head rotations~\protect\subref{fig:bosph_rot_f1}-\protect\subref{fig:bosph_rot_f2} and  incorrectly selected descent maps~\protect\subref{fig:bosph_dm_f1}-\protect\subref{fig:bosph_dm_f2}.}
\label{fig:bosph_samp}
\end{figure*}

\begin{figure}
\hfill\null
\subfloat[GRID]{\includegraphics[scale=.3]{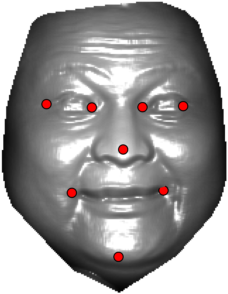}}
\hfill
\subfloat[SMUF]{\includegraphics[scale=.3]{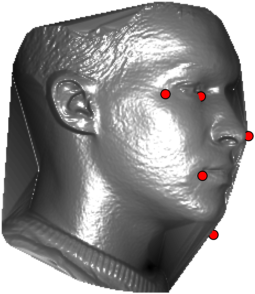}}
\hfill
\subfloat[GRID]{\includegraphics[scale=.3]{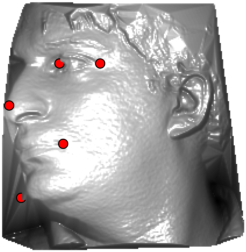}}
\hfill
\subfloat[SMUF]{\includegraphics[scale=.3]{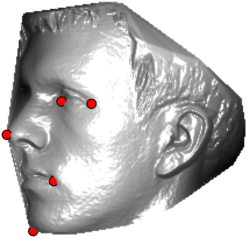}}
\hfill\null
\\
\null\hfill
\subfloat[GRID]{\includegraphics[scale=.3]{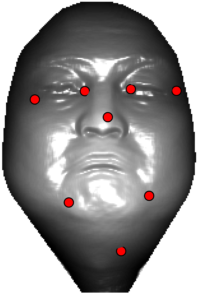}\label{fig:und_expr_f2}}
\hfill
\subfloat[SMUF]{\includegraphics[scale=.3]{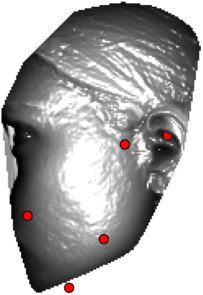}\label{fig:und_det_L45_f1}}
\hfill
\subfloat[GRID]{\includegraphics[scale=.3]{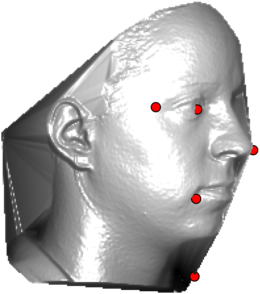}\label{fig:und_chin_R60_f1}}
\hfill
\subfloat[SMUF]{\includegraphics[scale=.3]{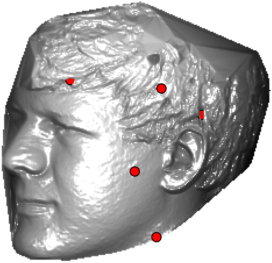}\label{fig:und_dm_L60_f1}}
\hfill\null
\caption{Exemplar landmark detection results on the UND and FRGCv2 datasets: the first row depicts test images with typical localization performance, while images from the second row are selected among the \textbf{worst} samples measured by the localization error.}
\label{fig:und_samp}
\end{figure}
\subsection{Landmark Analysis}

In this section we evaluate how the overall localization performance varies across the individual landmarks for both GRID and SMUF. Fig.~\ref{fig:lm_var} illustrates the mean localization errors achieved for the individual landmarks -  the size of the circles is proportional to the errors. It can be observed that the landmarks corresponding to the nose tip and eye and mouth corners exhibit low localization errors. This is expected as these landmarks correspond to well pronounced facial parts with distinctive ``corner-like'' shapes. Contrary, landmarks relating to nose saddle points, the chin tip and eyebrow points correspond to indistinctive ``edge-like'' local shapes and, therefore, result in high localization errors. This observation is also supported by the box plots in Figs.~\ref{fig:boxplot_bosph} and \ref{fig:boxplot_und} that show localizations errors of individual landmarks on the Bosphorus and the FRGCv2+UND databases.The presented behaviour is consistent for both evaluated methods.

To further analyze the landmarking performance of GRID and SMUF and their generalization ability, we also performed a cross-database experiment, where we built the test set with images from the FRGCv2+UND dataset, while the training set was generated using images from the Bosphorus dataset. Results relating to the cross-database experiment are illustrated by the green box plots in Fig.~\ref{fig:boxplot_und}. When compared to the experiment where both training and test sets are from the FRGCv2+UND dataset, we observe a slight increase in localization errors for most of the landmarks, except the chin tip where the difference between mean errors is larger. We presume that the high mean error for the chin tip comes from the increased appearance variability caused by the face detection procedure needed for the FRGCv2+UND data. (see Fig.~\ref{fig:und_chin_R60_f1}). Since such variability is not present in the training set from the Bosphorous dataset, the landmarking procedures cannot learn to accommodate for the inaccuracies of the face detector. In terms of comparison of GRID and SMUF, we see no significant difference in their performance in these experiments.

\subsection{Qualitative Evaluation}

In this section we qualitativelly assess the landmarking performance of the proposed landmarking methods. Figs.~\ref{fig:bosph_samp} and \ref{fig:und_samp} show exemplar face images from the Bosphorus and UND datasets with localized landmarks marked by red dots. The top rows of both figures contain samples with typical localization performance, where we can see that the method possess stable performance in the presence of different types of variability, such as expressions, partial occlusions and head rotations. However, there are some cases where landmarks are poorly localized. Such samples with large localization errors are exposed in the second rows of Figs.~\ref{fig:bosph_samp} and \ref{fig:und_samp}. E.g., large occlusions of face areas (Figs.~\ref{fig:bosph_occl_f1} and \ref{fig:bosph_occl_f2}) can cause increased localization errors of visible landmarks. Some of the localization errors originate from poor face detection and cropping, where an image can contain also non-head regions (Figs.~\ref{fig:und_expr_f2} and \ref{fig:und_chin_R60_f1}) or parts of the face area are cropped out (Fig.~\ref{fig:und_det_L45_f1}). Mis-selected descent maps can also be the cause of landmark localization errors (Figs.~\ref{fig:bosph_dm_f1}, \ref{fig:bosph_dm_f2} and \ref{fig:und_dm_L60_f1}).

\subsection{Computational Cost}

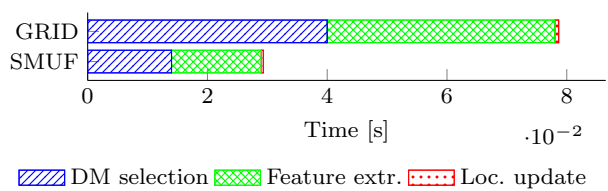
\begin{figure}
\centering
\begin{tikzpicture}
\begin{axis}[
xbar stacked,
legend style={
legend columns=4,
at={(xticklabel cs:-0.15,25)},
anchor=west,
cells={anchor=west}, 
draw=none
},
ytick=data,
axis y line*=none,
axis x line*=bottom,
tick label style={font=\footnotesize},
legend style={font=\footnotesize},
label style={font=\footnotesize},
xticklabel style={/pgf/number format/fixed},
width=1.0\linewidth,
bar width=3mm,
xlabel={Time [\si{\second}]},
yticklabels={SMUF,GRID},
xmin=0,
area legend,
y=4mm,
enlarge y limits={abs=0.625},
]
% DM selection
\addplot[blue,fill=none,postaction={pattern=north east lines,pattern color=blue}] coordinates{(0.014,0)(0.040,1)};
\addlegendentry[align=left]{DM selection}
% Feature extraction
\addplot[green,fill=none,postaction={pattern=crosshatch,pattern color=green}] coordinates{(0.015,0)(0.038,1)};
\addlegendentry[align=left]{Feature extr.}
% Updates
\addplot[red,fill=none,postaction={pattern=dots,pattern color=red}]coordinates{(0.00033,0)(0.00065,1)};
\addlegendentry[align=left]{Loc. update}
\end{axis}
\end{tikzpicture}
\vspace{0mm}
\caption{Average running time of the GRID and SMUF methods to localize landmarks on one face image (computed over 100 randomly selected test images). For the bencmarking images from the Bosphorous dataset were used and $22$ landmarks were predicted. The results show that SMUF is around $3 \times$ faster than GRID, but ensures only slightly higher localization errors.}\label{fig:time}
\end{figure}

In the last series of the experiments we evaluate the time needed by the GRID and SMUF methods to localize landmarks on a single test image on average. We compute the average processing time over $100$ randomly selected test images from the Bosphorous dataset. The size of the input images is $250 \times 200$ and we compute the locations of all $22$ landmarks during the benchmark. A PC with the following specifications is used for the assessment: Intel Xeon CPU \SI{2.67}{\giga\hertz} with \SI{12}{\giga\byte} RAM. Both landmarking techniques theare implemented using Matlab and could be further sped up if implemented with a compiled language such as C/C++. We start from detected and localized face regions and measure the time for feature extraction, DM selection and location updates, which take less than \SI{3e-2}{\second} for SMUF (see Fig.~\ref{fig:time}) and little less than \SI{8e-2}{\second} for GRID. When compared to handcrafted features, the learned binary features can be extracted almost 3 times faster than HOG features and 15 times faster than LBP features.

\section{Conclusion and Future Work}\label{sec:concl}

We have presented two approaches to facial landmark localization from 3D face images, GRID and SMUF, that are robust to rotations, facial expressions and partially also to occlusions. We proposed a gating mechanism that allowed us to incorporate multiple pose-specific landmarking models (based on HOG features) into the alignment procedure and also developed a simultaneous descent map and binary feature learning algorithm around the proposed gating mechanism. To assess performance we evaluated the developed landmarking techniques on three challenging datasets, containing 3D face images with large head rotations. Our results showed that the proposed solutions exhibit high robustness to different types of appearance variations and display competitive performance when compared to the state-of-the-art.

As part of our future work we plan to combine the proposed landmarking methods with face frontalization (or pose correction) procedures and incorporate all developed methods into pose-invariant 3D face recognition systems.

\begin{acknowledgements}
This research was supported in parts by the ARRS (Slovenian Research Agency) Research Program P2-0250 (B) Metrology and Biometric Systems, the ARRS Research Program P2-0214 (A) Computer Vision, and the RS-MIZ\v{S} and EU-ESRR funded GOSTOP.
\end{acknowledgements}

\section*{Compliance with ethical standards}
{\small\noindent\textbf{Conflict of interest} The authors declare that they have no conflict of interest.}

%\bibliographystyle{spmpsci}    % mathematics and physical sciences
%\bibliography{refs.bib}        % name your BibTeX data base

\end{document}